\documentclass[conference]{IEEEtran}
\IEEEoverridecommandlockouts
\usepackage{cite}
\usepackage{bm}
\usepackage{amsmath,amssymb,amsfonts}
\usepackage{algorithm}
\usepackage{algorithmic}
\usepackage{graphicx}
\usepackage{textcomp}
\usepackage{xcolor}
\usepackage{diagbox}
\usepackage{caption}
\usepackage{subcaption}

\def\BibTeX{{\rm B\kern-.05em{\sc i\kern-.025em b}\kern-.08em
    T\kern-.1667em\lower.7ex\hbox{E}\kern-.125emX}}

\begin{document}

\title{GraphDefense: Towards Robust Graph Convolutional Networks}

\author{
\IEEEauthorblockN{Xiaoyun Wang }
\IEEEauthorblockA{Department of Computer Science \\
University of California, Davis \\
Email: xiywang@ucdavis.edu}
\and
\IEEEauthorblockN{Xuanqing Liu }
\IEEEauthorblockA{Department of Computer Science \\
University of California, Los Angeles \\
Email: xqliu@cs.ucla.edu}
\and
\IEEEauthorblockN{Cho-Jui Hsieh}
\IEEEauthorblockA{Department of Computer Science \\
University of California, Los Angeles \\
Email: chohsieh@cs.ucla.edu}
}

\maketitle
\thispagestyle{empty}
\pagestyle{empty}



\newcommand{\final} {0}

\ifthenelse{\equal{\final}{1}} {\renewcommand{\XY }[1]{}}{}
\ifthenelse{\equal{\final}{1}} {\renewcommand{\XQ }[1]{}}{}
\ifthenelse{\equal{\final}{1}} {\renewcommand{\CH }[1]{}}{}

\begin{abstract} 
In this paper, we study the robustness of graph convolutional networks (GCNs). Despite the good performance of GCNs on graph semi-supervised learning tasks, previous works have shown that the original GCNs are very unstable to adversarial perturbations. In particular, we can observe a severe performance degradation by slightly changing the graph adjacency matrix or the features of a few nodes, making it unsuitable for security-critical applications. Inspired by the previous works on adversarial defense for deep neural networks, and especially adversarial training algorithm, we propose a method called GraphDefense to defend against the adversarial perturbations. 
In addition, for our defense method, we could still maintain semi-supervised learning settings, without a large label rate.
We also show that adversarial training in features is equivalent to adversarial training for edges with a small perturbation. 
Our experiments show that the proposed defense methods successfully increase the robustness of Graph Convolutional Networks. Furthermore, we show that with careful design, our proposed algorithm can scale to large graphs, such as Reddit dataset. 

\end{abstract}

\section{Introduction}

The GCN model takes in both feature matrix $X$ and the adjacency matrix $A$, the original model consists of two fully connected layers parameterized by $W^{(1)}$ and $W^{(2)}$, together with a final softmax layer to do the per-node classification. In specific, we can formulate the whole model as
\begin{equation}
\begin{aligned}
Z &= \text{softmax} \big( \hat A  \sigma (\hat A X W^{(1)})W^{(2)}\big)\\
  &= \text{softmax}\big(f(X, A)\big), 
\end{aligned}
\label{eq:gcn}
\end{equation}
where $A$ is the original adjacency matrix and $\hat{A}=\tilde{D}^{-1/2}\tilde{A}\tilde{D}^{-1/2}$, is the normalized adjacency matrix. $\tilde{A}=A+I$ is the original graph plus ``self-connection'' and $\tilde{D}=\text{diag}\{\tilde{A}\bm{1}\}$ is the degree matrix of each node. Although it looks tempting to try augmenting with more layers so that the information can be diffused to further nodes in deeper layers, experimental results in~\cite{DBLP:journals/corr/KipfW16} shows that a two layer network is the most effective setting. One limitation of the original GCN is that it directly aggregates the feature vector of a certain node with its neighboring nodes, also the optimization algorithm requires to do full batch gradient descent, this is very inefficient when the training dataset is very large. 

To deal with this problem, neighbourhood sampling method came out \cite{graphsage}. GraphSAGE samples  a fixed size of neighbours for each nodes, and aggregates sampled neighbourhood features then concatenates with it own feature. After that, they use mini-batch during training. In this way, the memory bottleneck caused by randomness access is solved, thus working with large scale datasets and fast training become possible. 
The aggregation process for each node $v$ would be written as 
\begin{equation}
\begin{aligned}
    h_{N(v)}^{k} \leftarrow AGGREGATE_k (\{ h_u ^{k-1}, \forall u \in N(v)\})
\end{aligned}
\end{equation}

Where $N(v)$ is the sampled fixed number of neighbour of node $v$, $k$ is depth, and $h$ is the features vector or aggregate features vectors of nodes. $AGGREGATE$ is an aggregate function, we will use mean aggregator with GCN setting during our experiments:
\begin{equation}
\begin{aligned}
MEAN(\{h_{v}^{k-1}\} \bigcup \{ h_{u}^{k-1}, \forall u \in N(v) \} ).
\end{aligned}
\end{equation}

\par
Despite that GCN and its variants are suitable to deal with graph data, recently people found that they are also prone to adversarial perturbations. It is worth noting that such perturbations are unlike random noises, instead, they are usually created dedicatedly by maximizing the loss metric. By convention, we call the people who create such adversarial perturbations as ``attacker'' and the side who apply the model to testset as ``user''. For example, suppose the user is doing per-node classification, then it would be reasonable for attacker to maximize the negative cross-entropy loss over a testing example. The overall idea of finding adversarial perturbation can be described as a constraint optimization problem as follows
\begin{equation}
    \label{eq:adversarial-attack-idea}
    \delta=\mathop{\arg\max}_{\delta\in\mathcal{S}} J\big(f(x+\delta), y\big),
\end{equation}
where $J(\hat{y}, y)$ is the loss function and $f(x)$ is our model. $\mathcal{S}$ is the constraint depending of the goals of attack, two common choices are $\{\delta|\|\delta\|_2\le r\}$ and $\{\delta|\|\delta\|_{\infty}\le r\}$, both of them aim at creating an invisible perturbation if $r$ is small enough. 
\par
For deep neural networks on image recognition task, there are several ways to solve Eq.~\eqref{eq:adversarial-attack-idea} efficiently. The simplest one is called fast gradient sign method (FGSM)~\cite{fgsm}, where we do one step gradient descent starting from origin, that is, $\delta=\eta\cdot\text{sign}\big(\nabla_{\delta}J\big(f(x+\delta), y\big)\big)$, here we need to choose a step size $\eta$ properly such that $\delta\in\mathcal{S}$. It is shown that although simple, this method is quite effective for finding an adversarial perturbation for images. Moreover, it is straightforward to improve FGSM method by running it iteratively, and that is essentially projected gradient descent (PGD) attack~\cite{pgd}. 
\par
As to adversarial defense methods, we can roughly divide them into two groups: first method is to inject noises to each layer during both training and testing time, and hope that the additive noise can ``cancel out'' the adversarial pattern, examples include random-self ensemble~\cite{Liu_2018_ECCV}; Second method is to augment the training set with adversarial data, this is also called adversarial training~\cite{pgd}. Generally, for adversarial defense in image domain, adversarial training (the latter one) is slightly better than noise injection (the former one). However, in terms of adversarial training on graph data, there are several challenges that impede us from directly applying it to graph domain:
\begin{itemize}
\item Low label rates for semi-supervised learning setting. 
\item Due to the semi-supervised learning nature of GCN, if the nodes been perturbed are in testing group, then adversarial training may not work: this is because propagating the gradients to the nodes been attacked may require to go through several nodes, but in plain GCN model, each node can only access its 2-hop neighbourhoods.
\item Inductive learning is even more difficult, it remains unknown whether adversarial training on certain graph can successfully generalize to other graphs.
\end{itemize}

\section{Related Work}
\label{related}
\subsection{Attacks and Defense on CNNs}
Adversarial examples of computer vision have been studied extensively. ~\cite{fgsm} discovered that deep neural networks are vulnerable to adversarial attacks---a carefully designed small perturbation can easily fool a neural network. 
Several algorithms have been proposed to generate adversarial examples for image classification tasks, including FGSM \cite{fgsm}, IFGSM~\cite{ifgsm}, C\&W attack~\cite{cw} and PGD attack~\cite{pgd}. 
In the black-box setting, it has also been reported that an attack algorithm can have high success rate using finite difference techniques  \cite{zoo,cheng2018query}, and several algorithms are recently proposed to reduce query numbers  \cite{limitequrry,querylimit2}. 

Adversarial training is a popular way for improving robustness. It's based on the idea of including adversarial examples in the training phase to make neural networks robust against those examples.
For instance, 
 \cite{pgd, X16_2015arXiv151103034H,X17_2016arXiv161101236K} generate and append adversarial examples found by attack algorithms to training dataset. 
 Other methods modifying structures of neural networks such as modifying ReLU activation layers and adding noises into images to original training dataset \cite{X40_2017arXiv170706728Z}, modifying softmax layers and then use prediction probability to train ``student'' networks \cite{X24_2015arXiv151104508P}. Adding noises to images and using random self-ensemble helps with defensing white box attacks \cite{Liu_2018_ECCV}. Dropping or adding edges to graphs could be viewed as mapping adding noises methods for images to graphs. 

 Most of the above-mentioned works are focusing on problems with continuous input space (such as images), directly applying these methods to Graph Convolutional Networks can only improves robustness marginally.
 
\subsection{Nodes classifications tasks with GCNs}
GCNs widely are used for node classification tasks, the original one is introduced in \cite{DBLP:journals/corr/KipfW16}, after that tons of works came out. From the large scale training aspect, sampling from just a few neighbourhoods is a standard way to scale the algorithm to big datasets. Different sampling methods are introduced with different papers, such as uniform sampling \cite{graphsage}, importance sampling \cite{fastgcn}, sampling from random walk through neighbours \cite{pinsage}.

Different structure of GCNs also have been explored, for example more layers \cite{deepGCN}, change ReLU to Leaky ReLU \cite{leakyRelu}. And variety of aggregate functions have also been apply to GCNs, such as max pooling, LSTM, and other different pooling methods \cite{graphsage} \cite{improveGCN} \cite{AAAI1816329_pooling}. Original GCN could be used as a kind of mean aggregator inside of GraphSAGE.

\subsection{Attacks and Defense on GCNs}
The wide applicability of GCNs motivates recent studies about their robustness. 
\cite{netattack,pmlr-v80-dai18b,Wu19} recently proposed algorithms to attack GCNs by changing existing nodes' links and features. 
\cite{netattack} developed an FGSM-based method that optimizes a surrogate model to choose the edges and features that should be manipulated. 
\cite{pmlr-v80-dai18b} 
proposed several attacking approaches including, gradient ascent, Genetic algorithm and reinforcement learning; \cite{pmlr-v80-dai18b} also showed experiments of using drop edges and adversarial training for defensing, and claimed that dropping edges is a cheap way for increasing robustness. 
\cite{Wu19} learned graphs from a continuous function for attacking, also claimed that deeper GCNs have better robustness. 
Recently more defense methods come out, besides adversarial training, \cite{advDefense} used graph encoder refining and adversarial contrasting learning, this paper explores robustness on both original GCN and GraphSAGE for small datasets, large graphs' robustness has not been discussed yet.

\section{Defense Framework}
In this paper, we propose a framework for adversarial training for graphs to increase the robustness of GCNs. 
We will first introduce the adversarial training framework for GCN, and then discuss how to scale it up to large graphs and  the connection between feature perturbation and graph perturbation in GCN adversarial training. 

\subsection{Framework}
Unlike previous defense work for CNN, GCN has some unique characteristics that will cause difficulties for improving robustness of GCNs. 
\begin{itemize}
    \item \textbf{ Low labeling rate}:  
For most cases, GCN is used for classification nodes in graphs with semi-supervised setting, with lower labelling rate than supervised learning. It will lead to a problem if we directly apply adversarial training on it. For example, when attacking a GCN, the perturbations of edges and features will be limited almost limited to training datasets and their neighbours. Directed attacks are more powerful than indirected ones \cite{netattack}. Thus during adversarial training only a few nodes get adversarial examples. For example a node in testing dataset may at least be the 4-hop neighbour of the training nodes. While GCNs are usually 2 layers or 3 layers, thus transfer adversarial training information will be impossible for that nodes, so adversarial training fail to work in this case. 
    \item \textbf{ Less of transferability for adversarial training }:
Consider depth $K$ GCN, the adjacency matrix $A$ is multiplied $K$ times, and each node could get information for k-hop neighbour, but as the result of matrix multiplications, the further nodes (more-hop nodes) have less influence. Thus after adversarial training, if a testing nodes that are far from all adversarial examples, it will be more vulnerable than the nodes in the training test or close the them. 
    
\end{itemize}

For most cases, GCN is used for classification nodes in graphs with semi-supervised setting, with lower labelling rate than supervised learning. It will lead to a problem if we directly apply adversarial training on it. For example, when attacking a GCN, the perturbations of edges and features will be limited almost limited to training datasets and their neighbours. Directed attacks are more powerful than indirected ones \cite{netattack}. Thus during adversarial training only a few nodes get adversarial examples. For example a node in testing dataset may at least be the 4-hop neighbour of the training nodes. While GCNs are usually 2 layers or 3 layers, thus transfer adversarial training information will be impossible for that nodes, so adversarial training fail to work in this case.

\textbf{Proposed algorithm.}
It has been reported in~\cite{pmlr-v80-dai18b} that directly applying existing methods can only marginally improve the robustness of GCN. Due to lack of connectivity between training set and tested nodes that are being attacked, i.e. (they are in different connected components or they are not directly connected), the loss gradient of training set hard to be transmit to targets nodes. That is because when multiply adjacency matrix in GCNs, the further nodes will have small values and closer ones will have larger values (It is similar to Katz Similarity.), Thus targets nodes that is not in the same connected components will not get benefit from the adversarial training. And these far away from the adversarial training set only benefit marginally from the adversarial training. (definition for distance is the shortest pass from the targets node to any node in the adversarial training set.) 

With small labeling weight for semi-supervised learning, lack of connectivity is very common.  \cite{deepGCN} shows that using part of predicted labels as training labels could increase the accuracy for prediction when label rate is low. This gives us intuition to relief the less of transferability problem during adversarial training.

Thus, we introduce the proposed adversarial training objective function as:
\begin{equation}
    \min_{W^{(1)}, W^{(2)}} \{
    {\max_{||A' - A|| < \epsilon }J (A' \sigma (A' X W^{(1)})W^{(2)}, y)}
    \},  
    \label{eq:minmax}
\end{equation}
where $A'$ is the modified adjacency matrix. For efficiency, we do not constraint elements of $A'$ to be discrete. 
The loss function $J$ is defined as 
\begin{align*}
    &J(A' , X) =
    \sum_{i \in L}  loss(y_i, z_i) + \alpha \sum_{j \in U}  loss(\hat{y_i}, z_i) \\
    &   
    loss(y_i,z_i) =  \left( \max\Big([f(X, A')]_{i,:}\Big) - [f(X, A')]_{i, y_i} \right), 
\end{align*}
where $L$ is labeled nodes set and $U$ is unlabeled nodes set. 
The loss of labeled data and unlabeled data are combined with a weight $\alpha$. Using fitted label for unlabeled data will resolve the connectivity problem. 
We use this method to give each nodes a label(the label maybe correct or incorrect), thus during the adversarial training, each node are able to be in the training set. 

\begin{algorithm}[tb]
\caption{Framework for adversarial training}
\label{alg:framework}
   \begin{algorithmic}
   \STATE \textbf{Input:} Graph adjacency matrix $A$ and features $X$, and classifier $f$ and adversarial generating function $adv$, and loss function $J$
   \STATE \textbf{Output:} Robust classifier $f'$.
   \STATE Predict nodes labels using GCN :  $\hat y \leftarrow  f(A,X)$
   \STATE \textbf{for :} t = 0 to T-1 \textbf{do}
   \STATE \hspace{\algorithmicindent} Randomly sample w group of nodes noted as $Nodes_{adv}$ and $Nodes_{clean}$;
   \STATE \hspace{\algorithmicindent} get adversarial examples of the group of $a$ nodes, $A' \leftarrow adv (A,X,f,Nodes_{adv})$
   \STATE \hspace{\algorithmicindent}  \textbf{retrain} $f$ with loss function $J = \sum_{i \in L}  loss(y_i, z_i)+ \alpha \sum_{j \in Nodes_{adv} \bigcup Nodes_{clean} }loss(\hat{y_i}, z_i) $; Note the retrained $f$ as $f'$
   \STATE \textbf{retrain} $f'$
   \end{algorithmic}
\end{algorithm}

There are different ways of getting adversarial examples: (1)  adversarial perturbation that constrained in the  discrete space. (2) the proposed GraphDefense perturbation in the continuous space. 

\begin{algorithm}[tb]
   \caption{adversarial training using discrete adjacency matrix }
   \label{alg:discrete}
\begin{algorithmic}
    \STATE {\bfseries Input:} Adjacency matrix $A$; feature matrix $X$; A classifier $f$ with loss function $loss = crossEntropy$; targeted nodes $Nodes_{adv}$
   \STATE {\bfseries Output:} adversarial example $A'$
   \STATE  {\bfseries Let} $e_{add}^* =(u^*,v^*) \leftarrow \arg \max  \nabla_{A} \sum_{i \in Nodes_{adv}} loss(\hat y_i, z_i) $ \\
   \STATE  {\bfseries Let} $e_{drop}^* =(u^*,v^*) \leftarrow \arg \min  \nabla_{A} \sum_{i \in Nodes_{adv}} loss(\hat y_i, z_i) $ \\
   \STATE {\bfseries if} $|\nabla_{A} [\sum_{i \in Nodes_{adv}} loss(\hat y_i, z_i) ]_{e^*_{add}}| > |\nabla_{A} [\sum_{i \in Nodes_{adv}} loss(\hat y_i, z_i)]_{e^*_{drop}}|$ :
   \STATE \hspace{\algorithmicindent} $A' \leftarrow  A + e^*_{add} $ 
   \STATE {\bfseries else}:
   \STATE \hspace{\algorithmicindent} $A' \leftarrow  A - e*_{drop} $ 
   \STATE \textbf{retrain:} $A'$

\end{algorithmic}
\end{algorithm}

Adversarial training and adversarial attacking are different situations for GCNs. During adversarial attacking the values of adjacency matrix and features are constraint on some certain space. For example if the adjacency matrix is normalized by row, them the sum of each after adversarial attacks on adjacency matrix should always be 1; if the adjacency matrix is discrete, adversarial attacks are not able to add a continuous weight edge (say 1.23) into the graph.
While for adversarial training, when generating adversarial example, there is not such a constraint, the values could be either discrete or continuous or even negative, which gives us a larger research space for adversarial examples. 

To solve the inner max problem in~\eqref{eq:minmax}, we use gradient descent on the adjacency matrix. The time complexity overhead compared to the original backpropagation   is $O(|V|^2)$ per update, where $|V|$ is the number of nodes.

\begin{algorithm}[tb]
   \caption{our algorithm: GraphDefense}
   \label{alg:continous}
\begin{algorithmic}
    \STATE {\bfseries Input:} Adjacency matrix $A$; feature matrix $X$; A classifier $f$ with loss function $loss$; targeted nodes $Nodes_{adv}$.
   \STATE {\bfseries Output:} adversarial example $A'$
   \STATE
   Compute $A' = \arg\max \sum_{i \in Nodes_{adv}} loss(\hat y_i, z_i) $
   \STATE $A' \leftarrow \beta A' + (1-\beta) A'^{T}$
  \STATE \textbf{retrain:} $A'$

\end{algorithmic}
\end{algorithm}

\subsection{Scaling to Large Datasets}
To scale up our attack and defense, we conduct experiments using GraphSage with GCN aggregator. 
The difficulties are:
\begin{itemize}
    \item For large GCN training with SGD, all the efficient  methods rely on sampled neighborhood expansion. Examples include GraphSage \cite{graphsage}, FastGCN \cite{fastgcn} and many others. 
    Unfortunately, currently there is no attack developed for the sampled neighborhood expansion process and it will introduce difficulty in backpropagation in adversarial training. 
   %
    \item Due to the large number of edges, the existing greedy methods are very time consuming. For example \cite{pmlr-v80-dai18b} \cite{meta_learning} needs to check the gradient values for all the edges at each iteration, which requires $O(|V|^2)$ time. 
    \item Due to the large number of nodes, adversarial edge changing examples in the adversarial training process, may not appear in the testing process. Thus the robustness will be affected. 
\end{itemize}

In our implementation we consider the neighborhood expansion used in GraphSAGE  with the GCN aggregator. The aggregator could be written as:
\begin{align}
   agg := A_1 \sigma(A_2 X W_1) W_2, 
\end{align}
where $\sigma$ is activation function; $A_1$ is sparse matrix containing neighborhood list : $B^1$ in Figure \cite{graphsage}, $A_2$ is a sparse matrix containing neighbor's neighborhood list $B^2$; other matrices are dense; we note predicted labels $\hat y = softmax(agg) $.

For large dataset adversarial training, we could still use the framework above, by only changing GCN function $f$ to GraphSAGE aggregator $agg$ and using mini-batch during training. 
The time complexity for each epoch is $O(B*B* N_1)$, where $N_1$ is number of sampled 1-hop neighbours. 

\subsection{Adversarial training in features}
\label{adv_in_feat}
For large scale graph convolutional networks, neighborhood sampling is a common way to scale up to large graphs. The basic idea to aggregate features of 1-hop and 2-hop neighbour then doing nodes classification. This gives us an intuition for doing adversarial training faster and for large-scale graphs. 
We could generate adversarial features and using these features for adversarial training. We could prove that any small perturbation in discrete edge space are all included in features perturbations in continuous space. The time complexity for retraining features is in each batch $O(|B* N_{features}|)$. Adversarial training on features will speedup adversarial process especially for large batch training, furthermore GCNs will also be more robust on edge perturbations. 
When considering modifying feature matrix $X$ with $\delta$ perturbation, the formula of GCN in Eq \ref{eq:gcn} will be: 
\begin{equation}
\begin{aligned}
Z &= \text{softmax} \big( \hat A  \sigma (\hat A (X+ \delta) W^{(1)})W^{(2)}\big)\\
  &= \text{softmax}\big( \hat A  \sigma (\hat A X W^{(1)} + \hat A \delta W^{(1)})W^{(2)}\big)\
\end{aligned}
\label{eq:feat_pert}
\end{equation}

For $\epsilon$ perturbation on graph $A$,  the formula of GCN in Eq \ref{eq:gcn} will be: 
\begin{equation}
\begin{aligned}
Z &= \text{softmax} \big( \hat (A+ \epsilon)  \sigma (\hat (A+ \epsilon) X W^{(1)})W^{(2)}\big)\\
\end{aligned}
\label{eq:edge_pert}
\end{equation}

Consider surrogate models without activation functions, 
\begin{equation}
\begin{aligned}
\delta = \hat{A}^{-1} [\hat{A}^{-1} (\hat{A} +\epsilon) (\hat{A} +\epsilon) X - \hat{A} X]
\end{aligned}
\label{eq:delta}
\end{equation}

\section{Experiments}
We use Cora, Citeseer, and Reddit attribute graphs as benchmarks. For Cora and Citeseer, we split the data into 15\% for training, 35\% for validation, and 50\% for testing; For Reddit dataset, we use the same setting as GraphSAGE paper, which is 65 \% for training 11 \% for validation and 24 \% for testing. Dataset descriptions could be find in Table \ref{tab:dataset}
\begin{table}[htbp]
    \caption{Description for Cora Citeseer and Reddits}
    \centering
    \resizebox{0.46\textwidth}{!}{\begin{tabular} {|c|c|c|c|c|c|} 
        \hline
        \textbf{Datasets}& \textbf{Nodes} & \textbf{Edges} & \textbf{Features} & \textbf{Classes} & \textbf{Features Type}\\
        \hline
        Cora & 2,708 & 5,429&7 &1,433 & discrete\\
        \hline
        Citeseer & 3,312 & 4,732&6 &3,703 & discrete\\
        \hline
        Reddits & 232,965 & 11,606,919&41& 602 & continuous\\
        \hline
        
    \end{tabular}}
    \label{tab:dataset}
\end{table}
We conduct experiments on both single node and a group of nodes. 

\subsection{Defense for GCN}
We conduct experiments to test the robustness of GCNs with different retrain GCN : drop edges, naive adversarial training in Algorithm \ref{alg:discrete}  and our method GraphDefense in Algorithm \ref{alg:continous} with our framework in Algorithm \ref{alg:framework}. 
Drop edges training is a cheap way to increase the robustness of GCN; Retraining with adversarial samples also works for defense attacks \cite{pmlr-v80-dai18b}. Our GraphDefense method gets the best results among these methods when defending adversarial defense in most cases. 

Table \ref{tab:gcn_defense} shows defense a 100 nodes group defense using different methods, for Cora dataset, the number of changed edges is 100 for each group of 100 nodes; for Citeseer dataset, due to the graph density is lower than Cora dataset, we chose to modify 70 edges for each group of 100 nodes. Our method successfully beats naive adversarial training and dropping edges, and increases the accuracy of GCNs for  more than  60 \% without changing the semi-supervised learning setting. 
\begin{table}
    \centering
    \caption{Result of average accuracy with different defense methods on GCN for attacking a group of 100 nodes}

    \resizebox{0.46\textwidth}{!}{\begin{tabular} {|c|c|c|c|c|} 
        \hline
        \textbf{Dataset}& \multicolumn{2}{|c|}{\textbf{Cora}} &  \multicolumn{2}{|c|} {\textbf{Citeseer}} \\
        \hline
        & before attack & after attack & before attack & after attack\\
        \hline
        Clean &0.8449 & 0.408 & 0.7434& 0.396 \\
        \hline
        drop edges &0.8338& 0.474 & 0.7409& 0.410 \\
        \hline
        discrete adversarial training A & 0.8301 &0.492 &0.7385& 0.404\\
        \hline
        Our method &0.8486 & 0.692 &0.7409& 0.628\\
    \hline  
    \end{tabular}}
    \label{tab:gcn_defense}
\end{table}
\begin{table}
    \centering
     \caption{Result of average accuracy with different defense methods on GCN for attacking singe nodes}
    \resizebox{0.46\textwidth}{!}{\begin{tabular} {|c|c|c|c|c|} 
         \hline
        \textbf{Dataset}& \multicolumn{2}{|c|}{\textbf{Cora}} &  \multicolumn{2}{|c|} {\textbf{Citeseer}} \\
        \hline
        & before attack & after attack & before attack & after attack\\
        \hline
        Clean &0.8449 & 0.370 & 0.7434& 0.440 \\
        \hline
        drop edges &0.8338& 0.374 & 0.7409& 0.452 \\
        \hline
        discrete adversarial training A & 0.8301 &0.554 &0.7385& 0.552\\
        \hline
        Our method &0.8486 & 0.540 &0.7409& 0.632\\
    \hline  
    \end{tabular}}
    \label{tab:gcn_1by1}
\end{table}
When attacking singe nodes, for each targeted node, we modify 1 edge in the graph. Table \ref{tab:gcn_1by1} shows the result for single node attacks by only dropping edges, adding edges or both. We notice that in both Table \ref{tab:gcn_defense} and Table \ref{tab:gcn_1by1} dropping edges method is the least robustness expect for original GCN. That is because adding edges are more efficient when attacking GCN, thus although dropping edges is a very fast way, the improvement of robustness is not significant compared with other methods. We also notice that for the Cora dataset, during single node attacks, discrete adversarial training is better than GraphDefense, the reason might be discrete adversarial training is more suitable for single node attacks. We will discuss this in the latter part. 

\begin{figure}[!tbp]
    \includegraphics[width=0.46\textwidth]{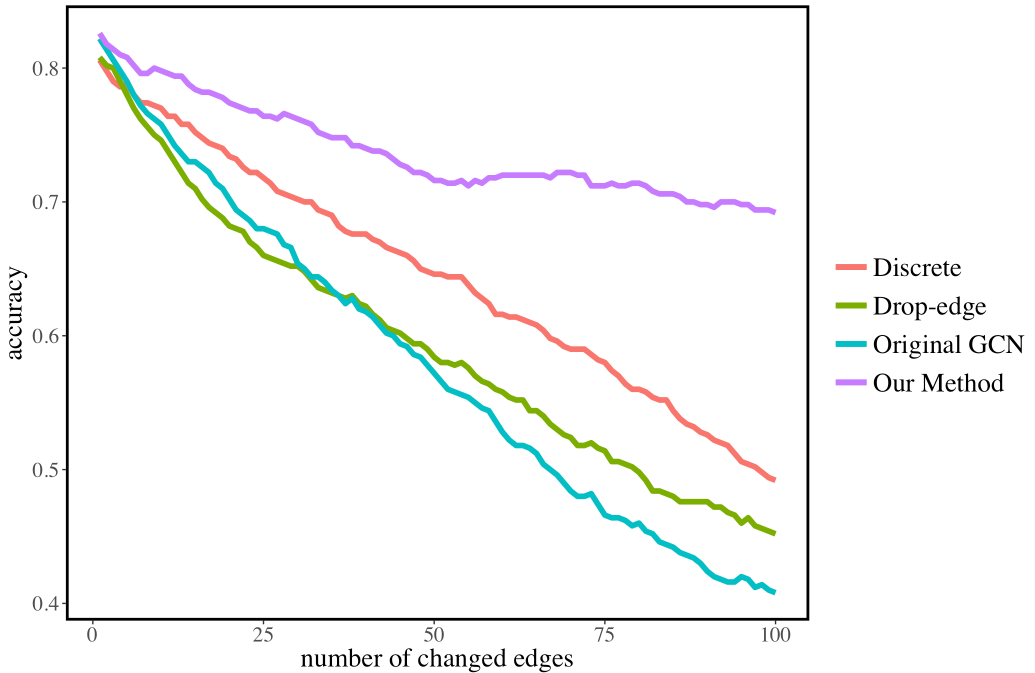}
    \caption{Accuracy of GCN under different defense methods for Cora, with modifying 0 to 100 edges. }
    \label{fig:cora_comp}
\end{figure}

\begin{figure}[!tbp]
    \includegraphics[width=0.46\textwidth]{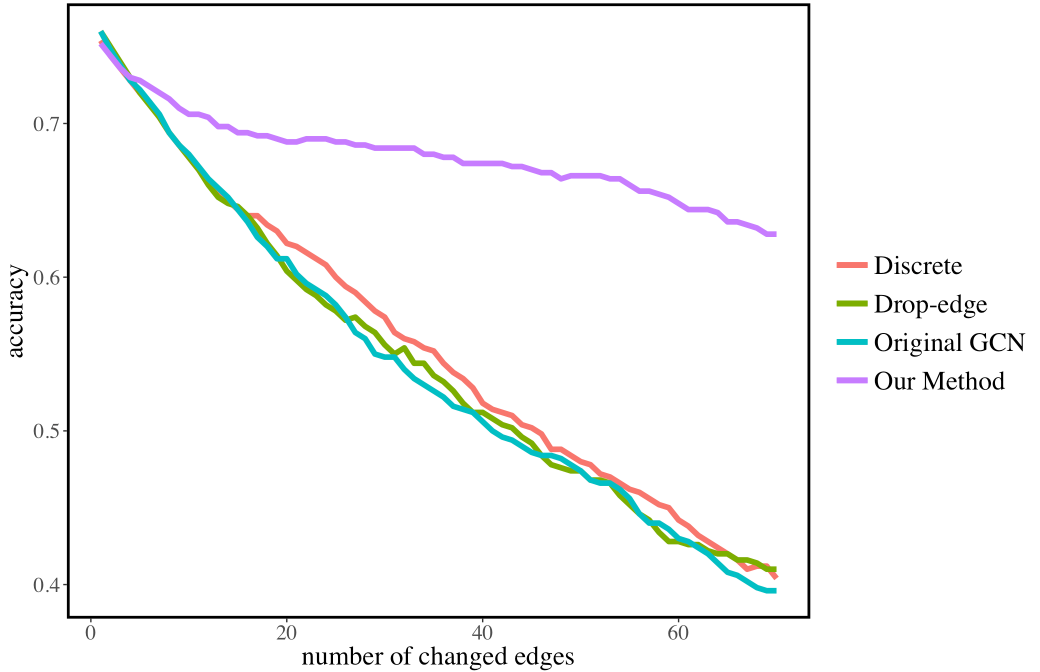}
    \caption{Accuracy of GCN under different defense methods for Citeseer, with modifying 0 to 70 edges. }
    \label{fig:citeseer_comp}
\end{figure}

Figure \ref{fig:cora_comp} and Figure \ref{fig:citeseer_comp} shows more details of attacking with different amount of modified edges. With the number of modified edges increases, Our GraphDefense method remains more stable than discrete adversarial samples retrain and dropping edges. To investigate deeper in the reason why these methods perform differently, we use to study the different degrees of nodes accuracy corresponding to attacks.

\begin{figure}[!tbp]
  \centering
  \includegraphics[width= 0.46\textwidth]{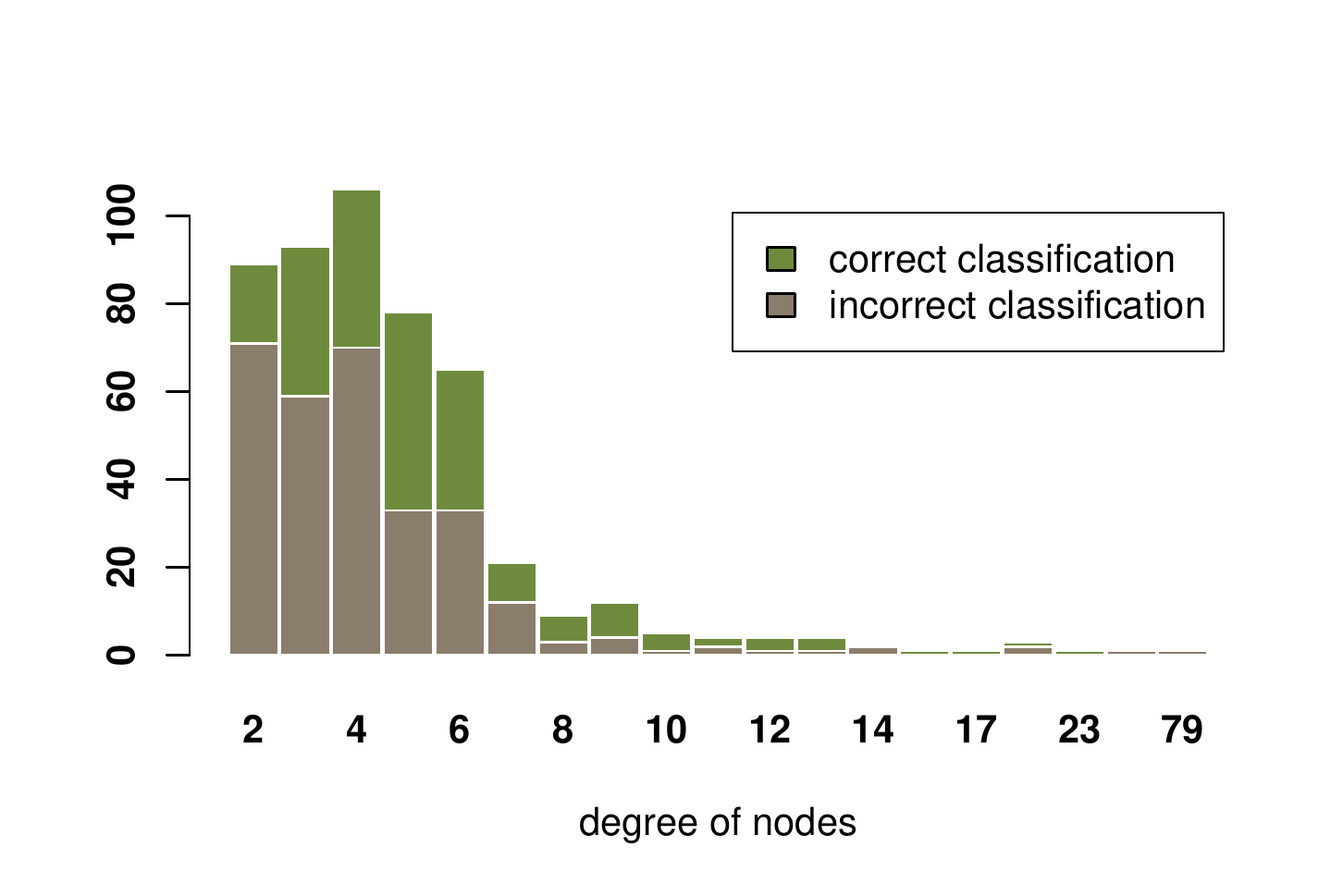}
    \caption{Classification performance, after attack for Original GCN, with different degrees of nodes}
    \label{fig:clean_deg}
\end{figure}
 \begin{figure}
     \centering
    \includegraphics[width=0.46\textwidth]{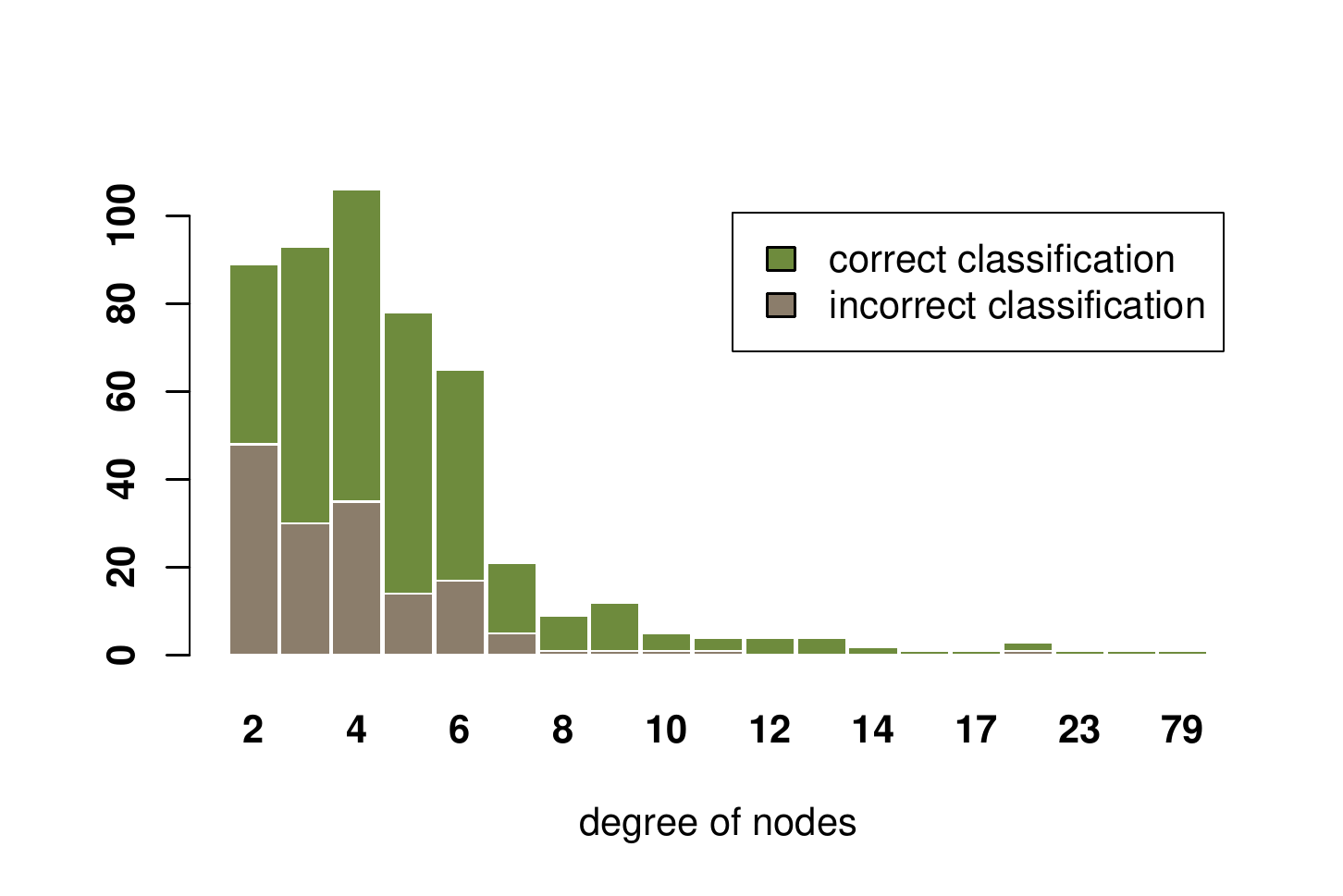}
     \caption{{Classification performance, after attack for our GraphDefense method, with different degrees of nodes}}
     \label{fig:conti_deg}
 \end{figure}

\begin{figure}
    \centering
    \includegraphics[width= 0.46\textwidth]{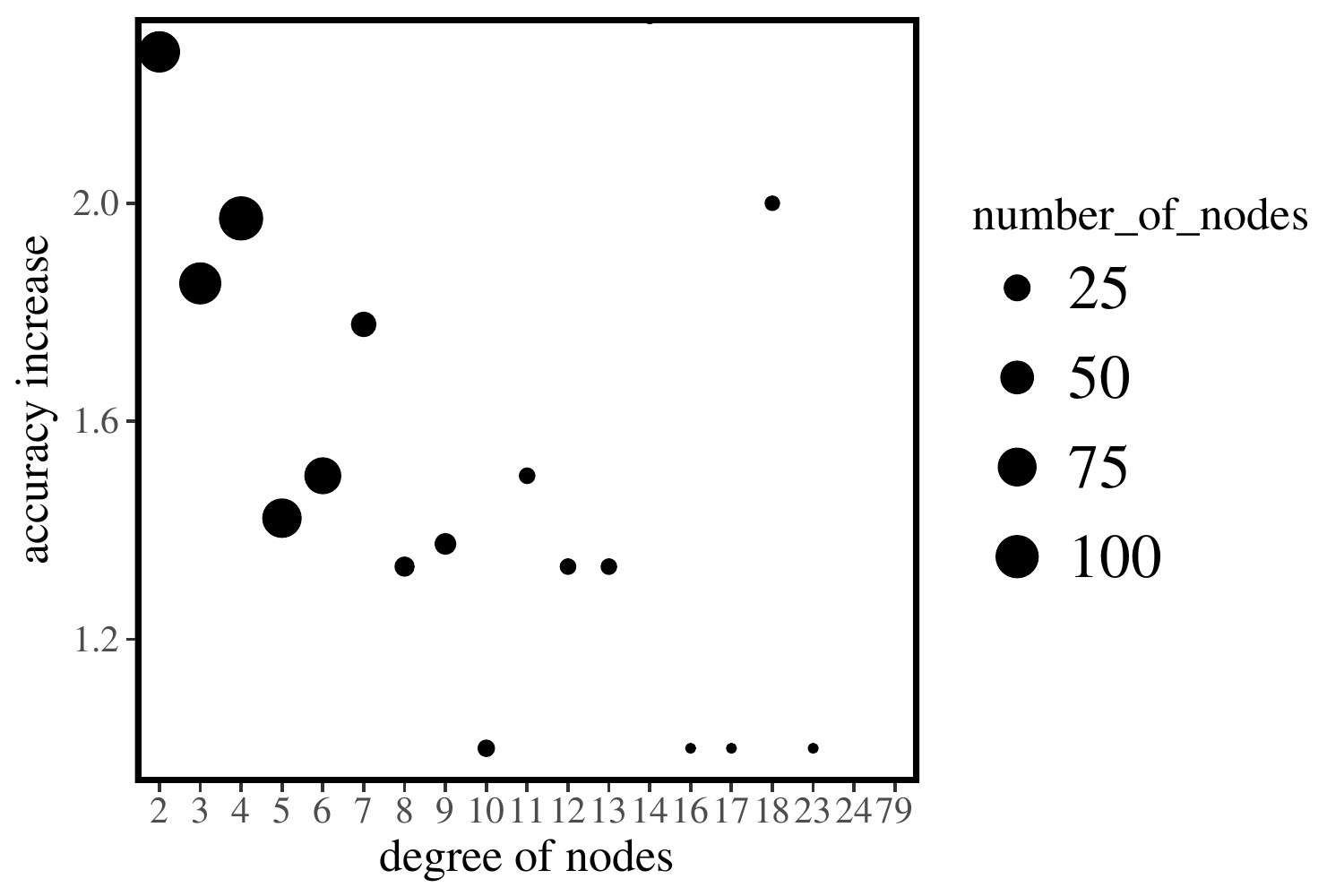}
    \caption{accuracy increase for nodes' degree, compare with  our defense method for Cora dataset; X axis: different degree of nodes; Y axis: accuracy improvement times}
    \label{fig:deg_cora}
\end{figure}

\begin{figure}
    \centering
    \includegraphics[width= 0.46\textwidth]{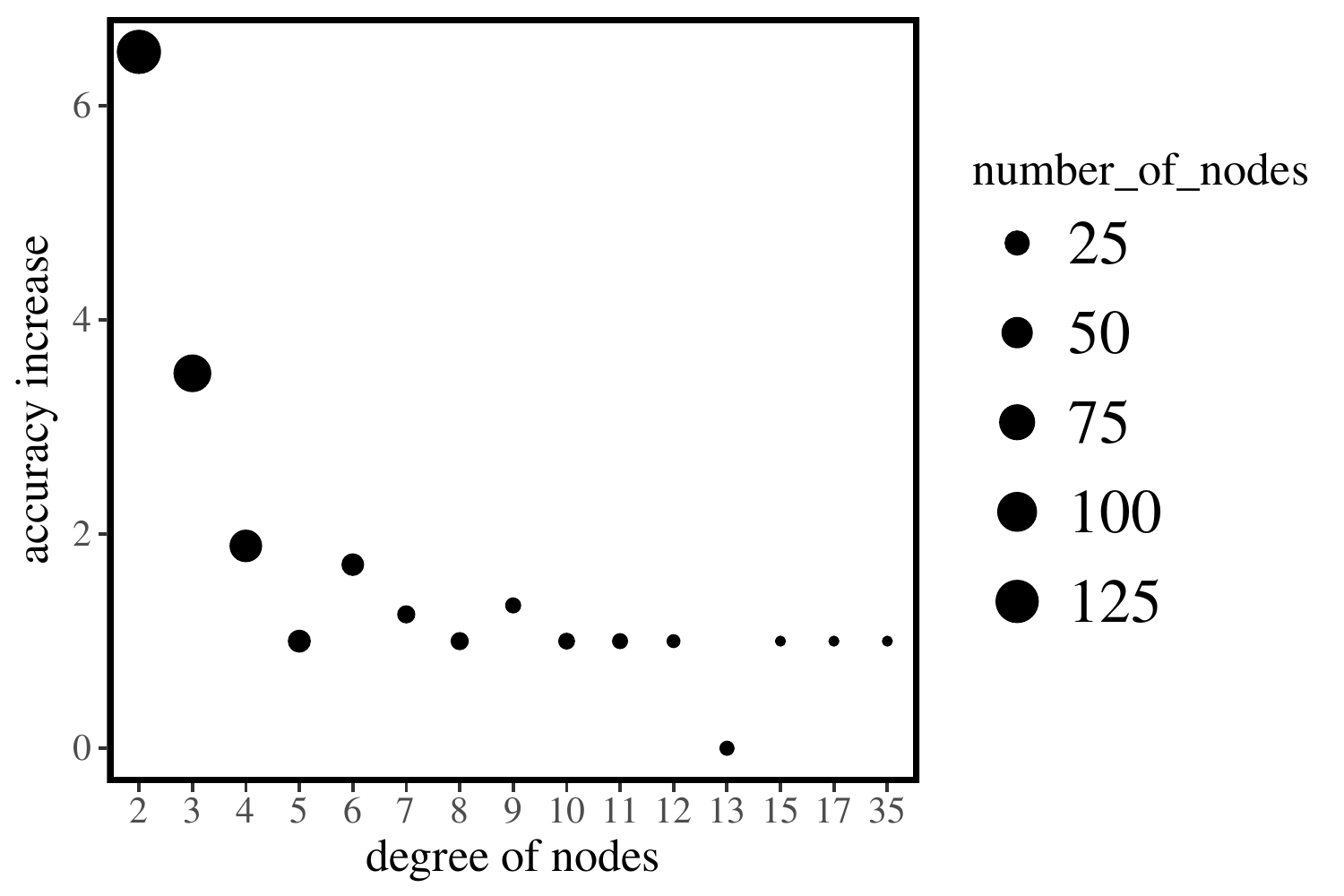}
    \caption {accuracy increase for nodes' degree, compare with  our defense method for Citeseer dataset; X axis: different degree of nodes; Y axis: accuracy improvement times}
    \label{fig:deg_citeseer}

\end{figure}

Figure \ref{fig:clean_deg} shows the correctly predicted nodes and incorrectly predicted nodes with the original GCN. It indicates that the lower degree nodes are more vulnerable. Figure \ref{fig:conti_deg} show accuracy ratio after attack with our GraphDefense method. The accuracy increases a lot for lower degree nodes. With degree follows power law distribution for most graph increasing lower degree nodes robustness is crucial for keep robustness of the GCNs. 
For Cora and Citeseer datasets, our graphDefense method works well for improving lower degree nodes robustness. Figure \ref{fig:deg_cora} and Figure \ref{fig:deg_citeseer} shows the accuracy improvement when compared with original GCNs. In the most case, our method gives a lower degree of nodes a boost on robustness after attacks. 

\begin{figure*}[!tbp]
  \centering
  \begin{subfigure}[b]{0.3\textwidth}
    \includegraphics[width=\textwidth]{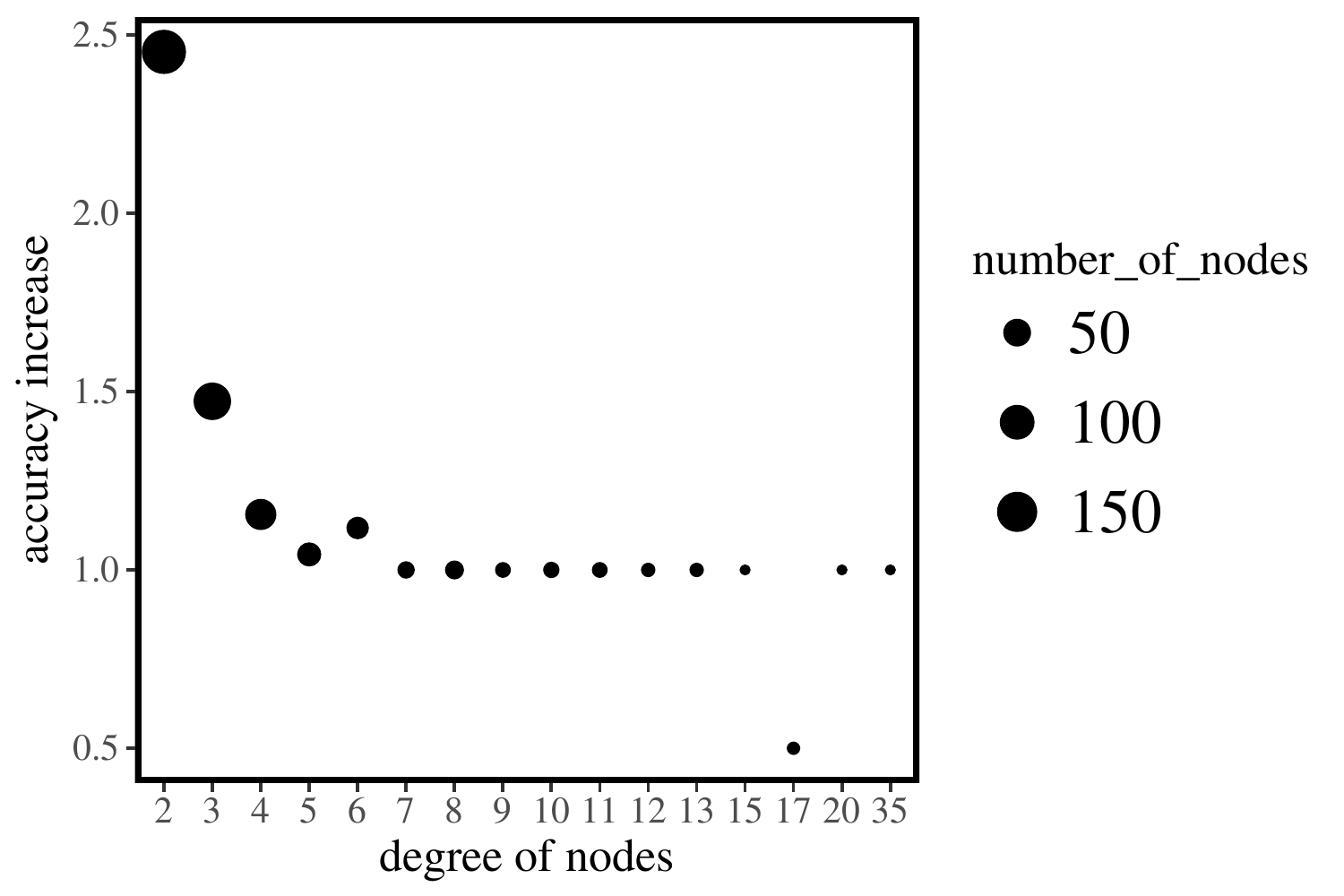}
    \caption{accuracy improvement times for nodes' degree, compare with our defense method and original GCN for attacking single node}
    \label{fig:1by1degree_conti_cite}
  \end{subfigure}
  \hfill
  \begin{subfigure}[b]{0.3\textwidth}
     \includegraphics[width=\textwidth]{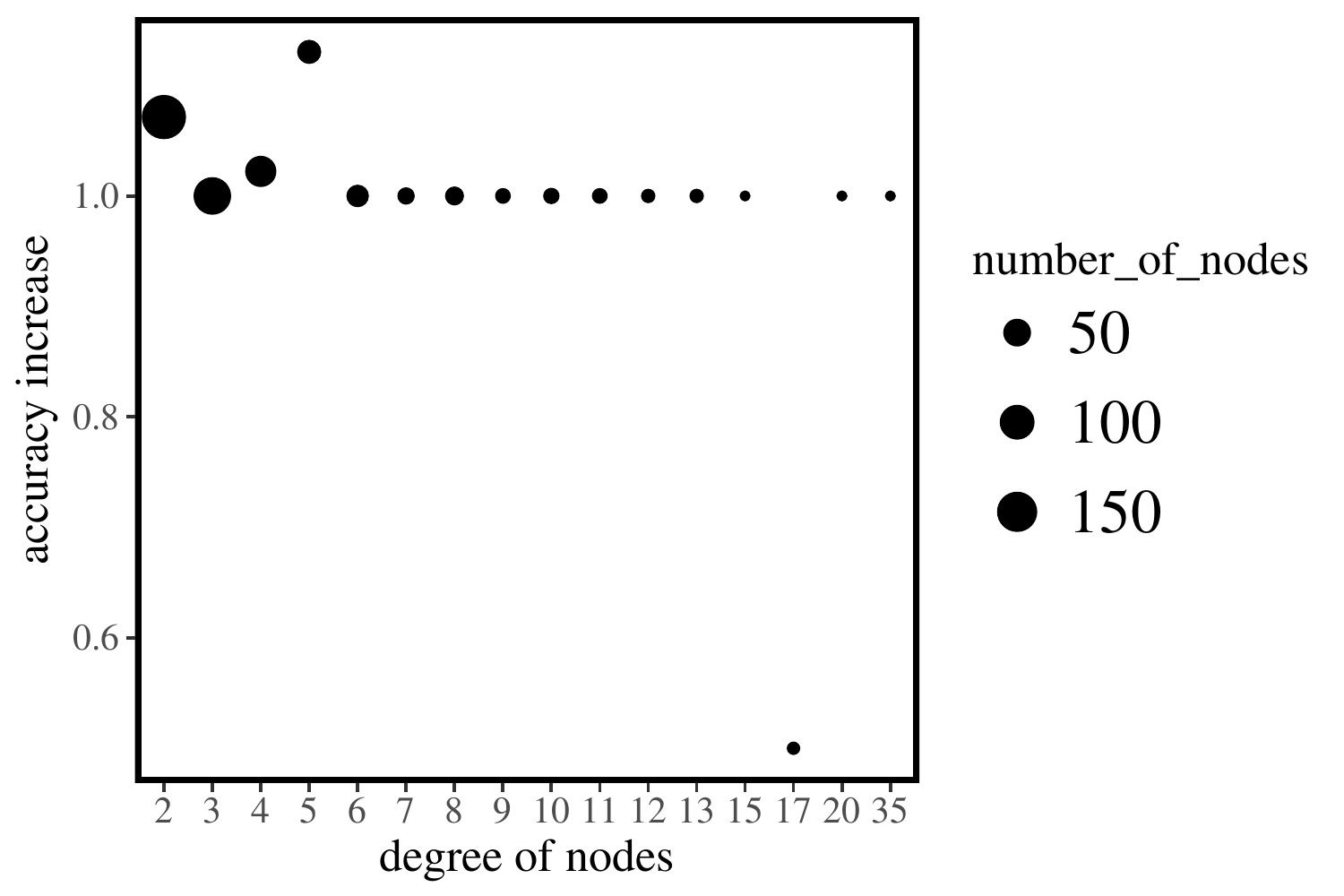}
    \caption {accuracy increase for nodes' degree, compare with dropping edges for attacking single node}
    \label{fig:1by1degree_drop_cite}
  \end{subfigure}
  \hfill
  \begin{subfigure}[b]{0.3\textwidth}
    \includegraphics[width=\textwidth]{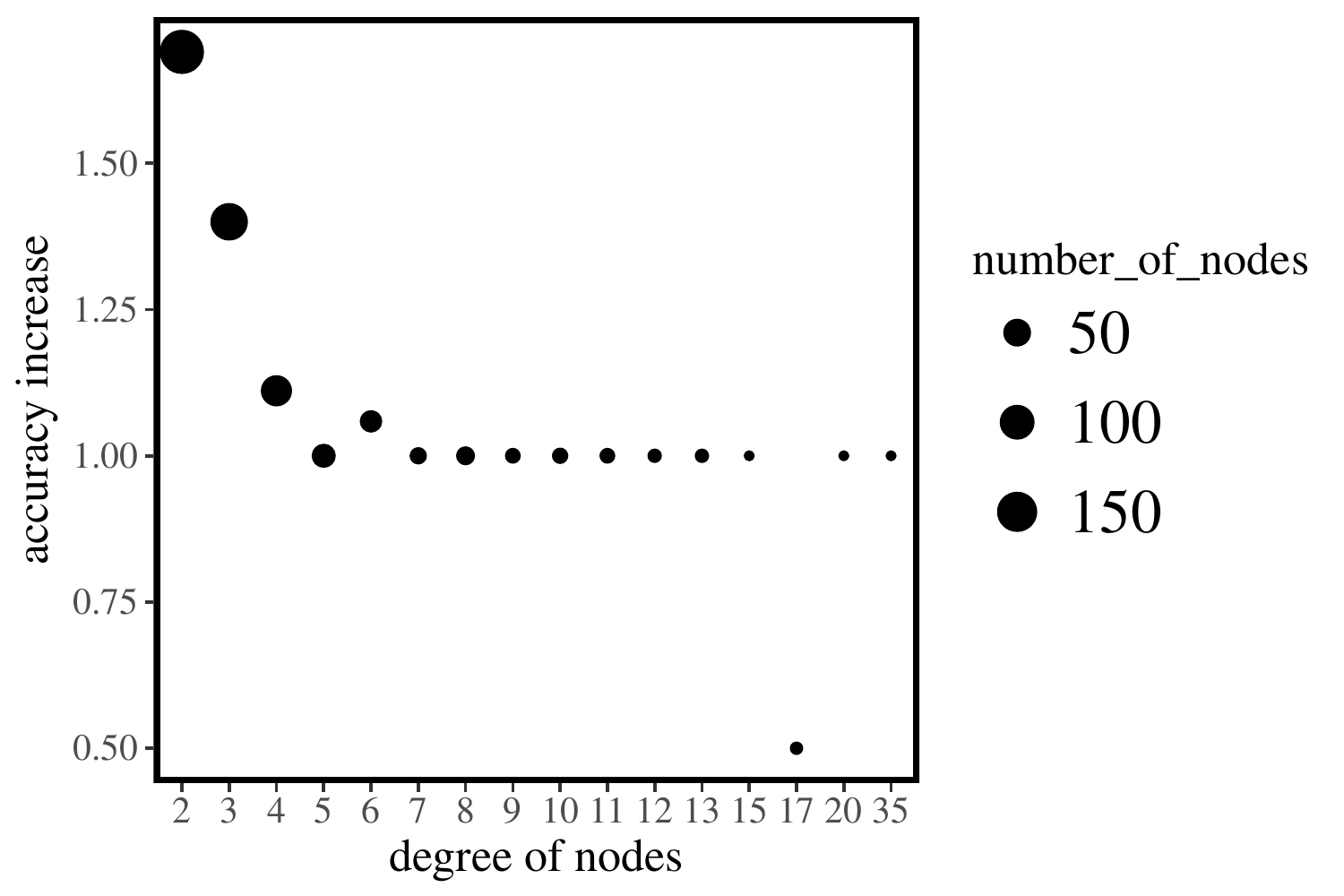}
    \caption {increasing accuracy rate for attacking single node by each degree for our method for attacking single node}
    \label{fig:1by1degree_disc_cite}
  \end{subfigure}
  \hfill
  \begin{subfigure}[b]{0.3\textwidth}
    \includegraphics[width=\textwidth]{figs/degree_conti_cite.pdf}
    \caption {increasing accuracy rate for attacking groups of 100 nodes by each degree for our method}
    \label{fig:degree_conti_cite}
  \end{subfigure}
  \hfill
  \begin{subfigure}[b]{0.3\textwidth}
    \includegraphics[width=\textwidth]{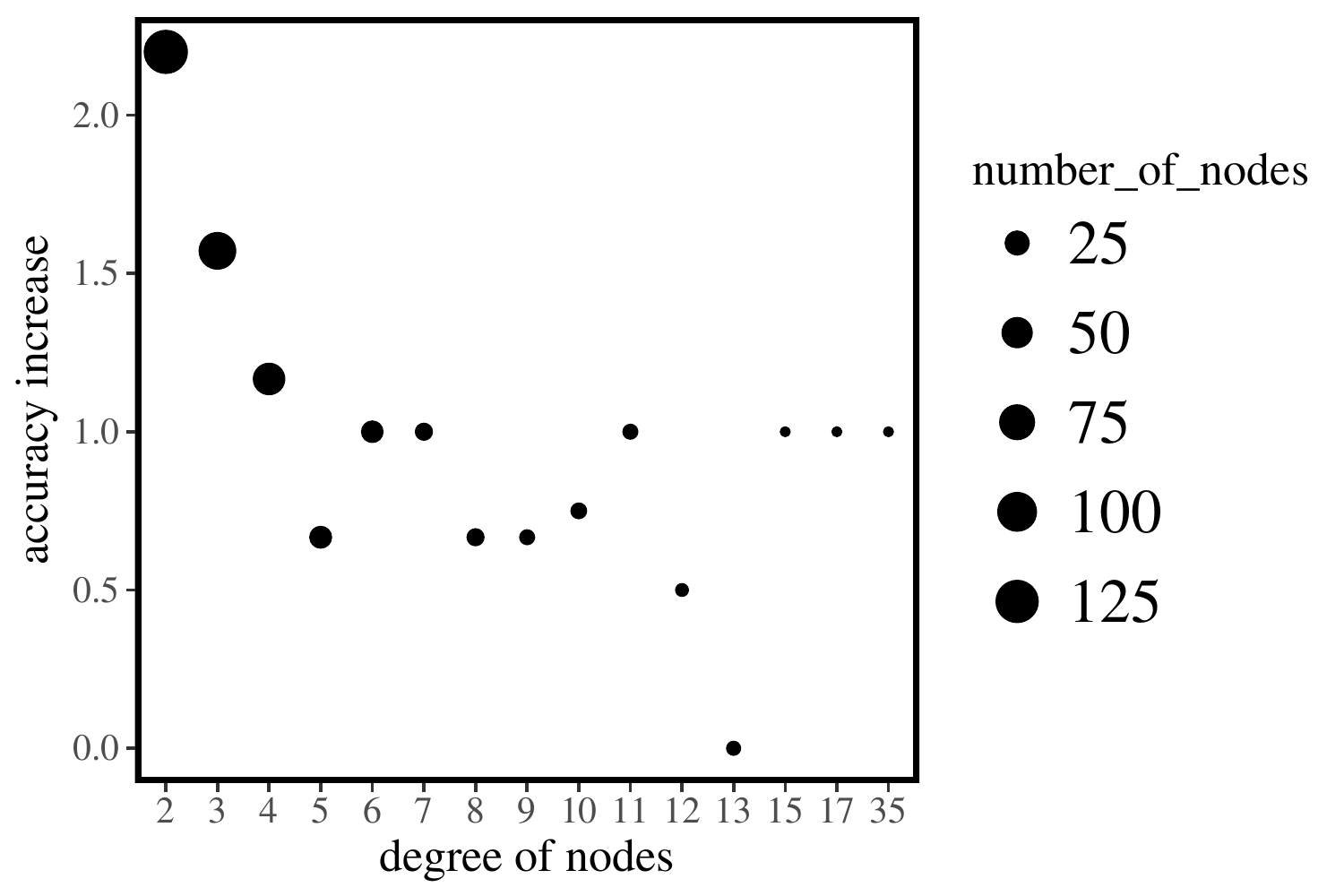}
    \caption {increasing accuracy rate for attacking groups of 100 nodes by each degree for our method}
    \label{fig:degree_drop_edges_cite}
  \end{subfigure}
  \hfill
  \begin{subfigure}[b]{0.3\textwidth}
    \includegraphics[width=\textwidth]{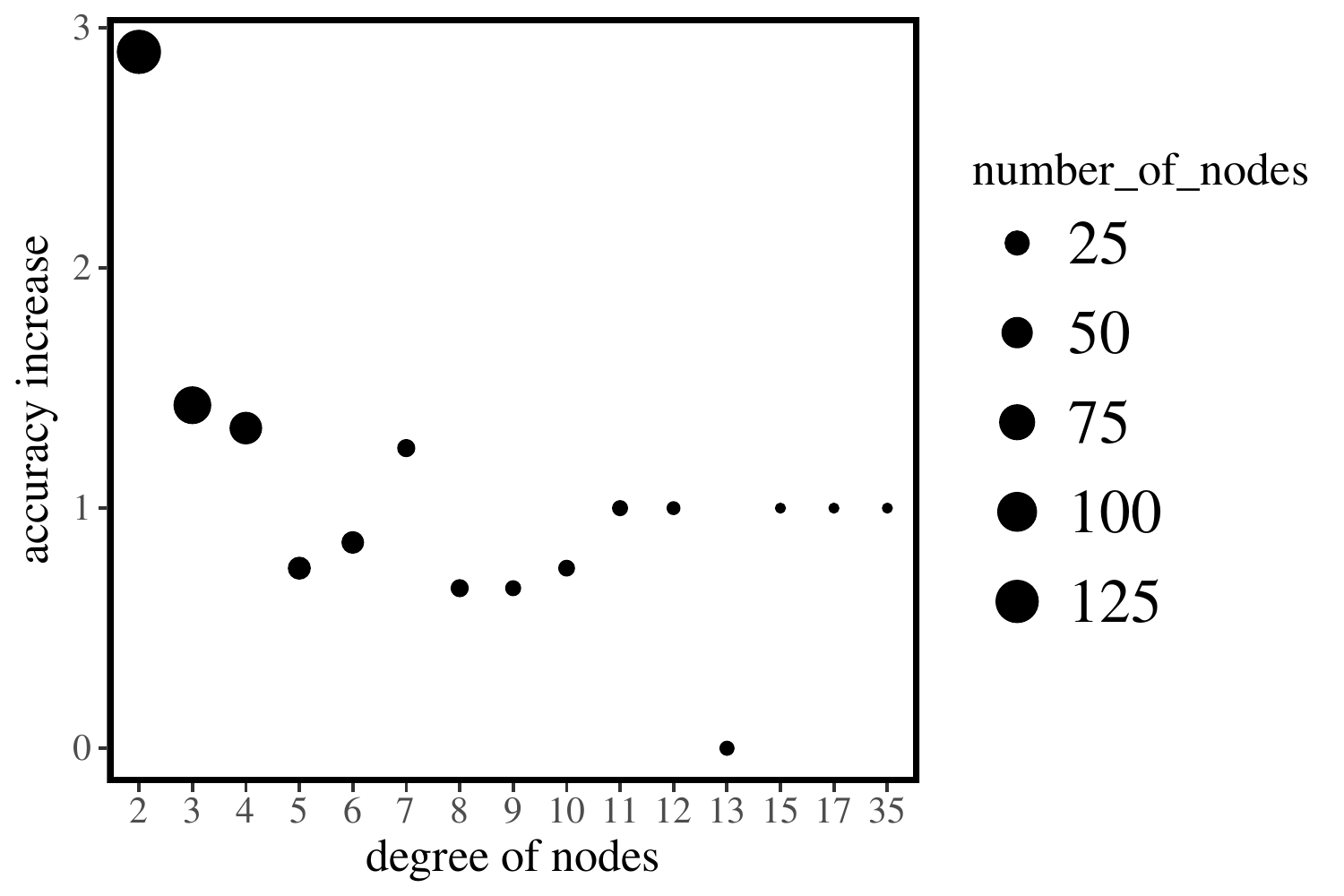}
    \caption {increasing accuracy rate for attacking 100 nodes by each degree for our method}
   \label{fig:degree_disc_cite}
  \end{subfigure}
\caption{Accuracy improvement for different attacks using discrete adversarial training, drop edges training and our GraphDefense method.}
\end{figure*}

Next, we are discussing how accuracy increasing for different methods. We use the Citeseer dataset as an example. Figure \ref{fig:1by1degree_conti_cite} \ref{fig:1by1degree_drop_cite} \ref{fig:1by1degree_disc_cite} show accuracy improvement for single node attacks, and Figure \ref{fig:degree_conti_cite} \ref{fig:degree_drop_edges_cite} \ref{fig:degree_disc_cite} is for attacking groups of 100 nodes. Our method not only keeps the higher degree nodes accuracy but also boost the lower degree ones. When comparing attacking groups of nodes and attacking single node, we find there our GraphDefense method results stay inconstant for different kinds of attacks, and the accuracy for degree 2 nodes improved by 6X for attacking groups nodes compared with 3X for attacking single node. While for the other 2 methods, the accuracy drops in some larger degree nodes for attacking groups of 100 nodes. 

\begin{figure*}[!tbp]
  \centering
  \begin{subfigure}[b]{0.3\textwidth}
    \includegraphics[width=\textwidth]{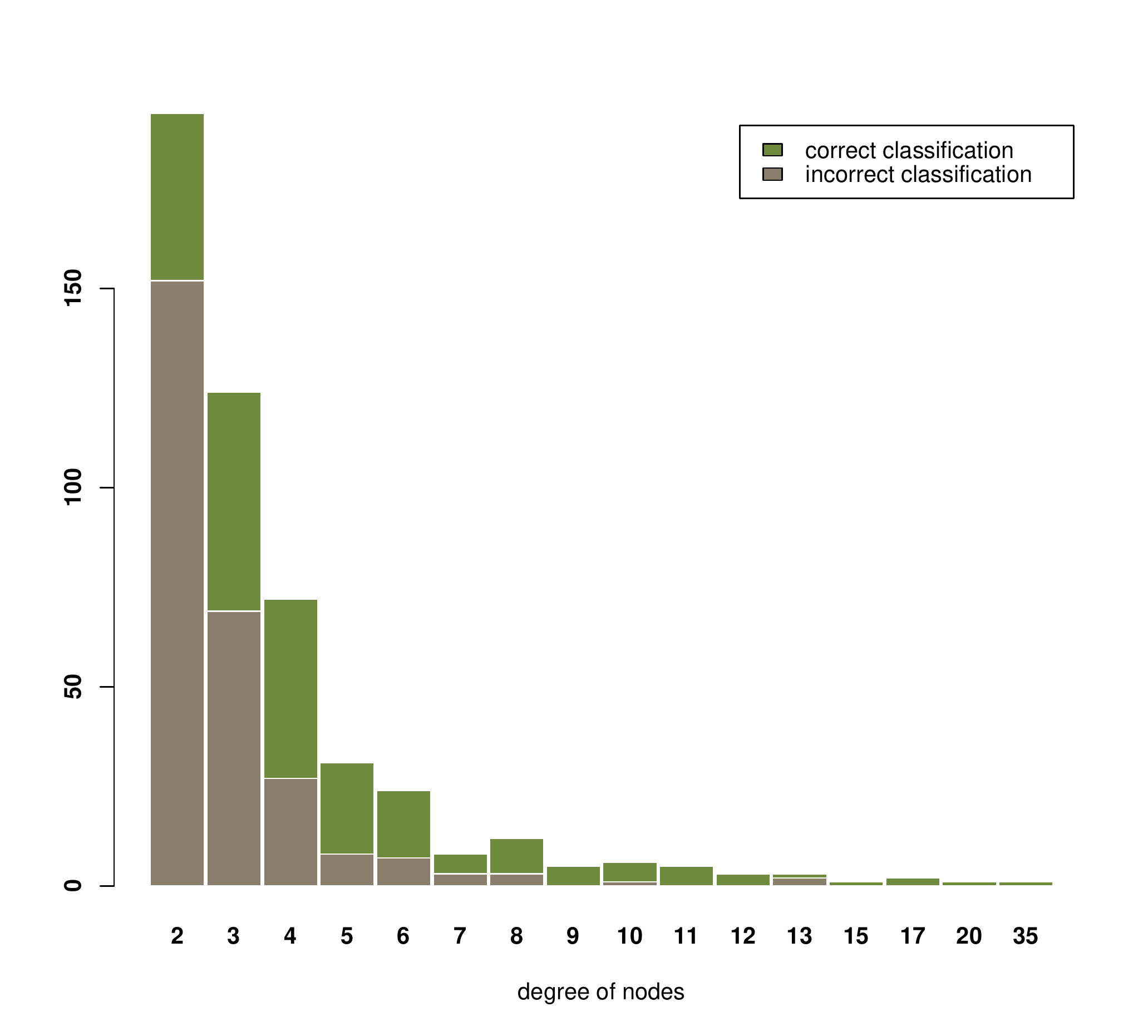}
    \caption{attack single nodes, Citeseer dataset, clean model.}
    \label{fig:1by1bar_gcn_cite}
  \end{subfigure}
  \hfill
  \begin{subfigure}[b]{0.3\textwidth}
    \includegraphics[width=\textwidth]{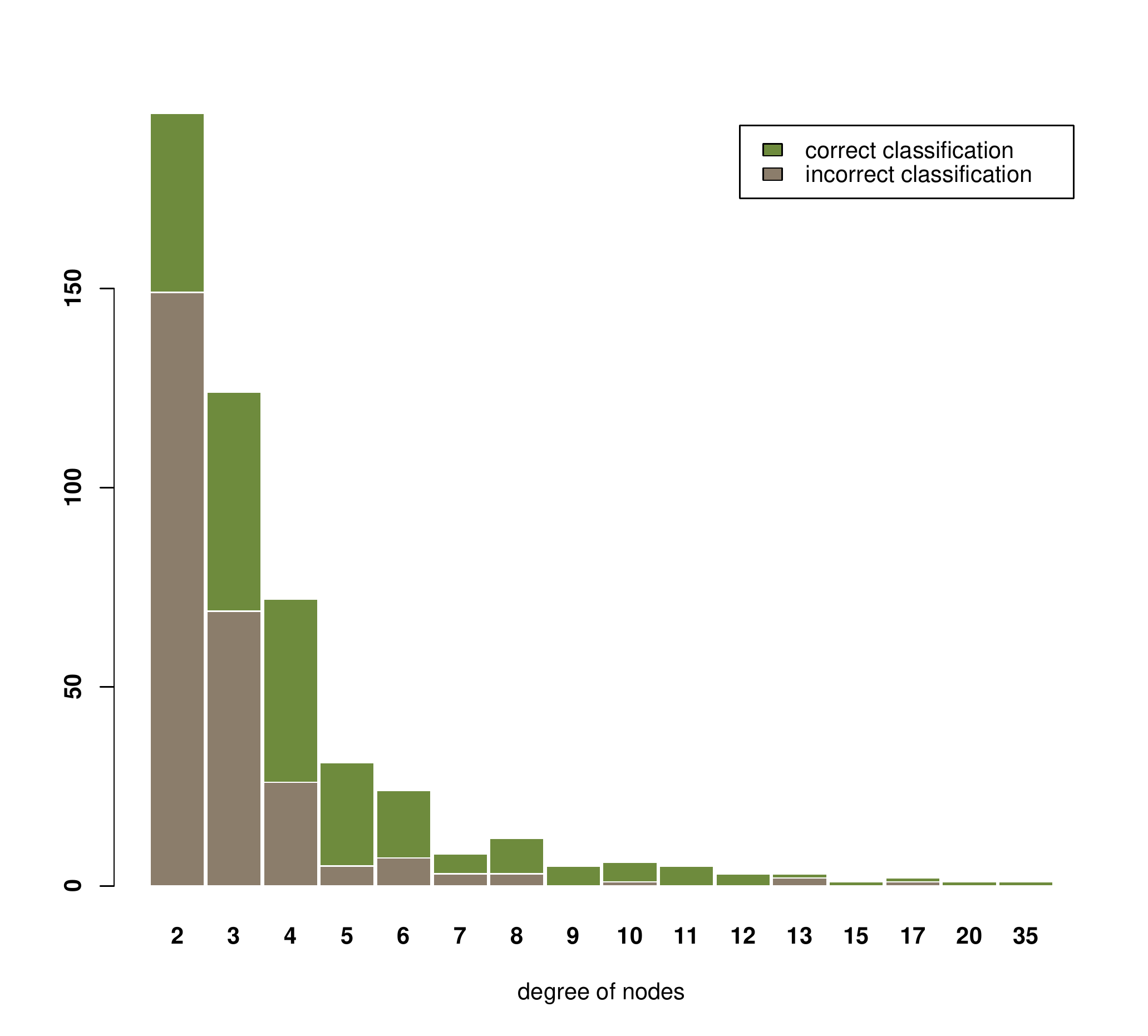}
    \caption{attack single nodes, Citeseer dataset, using drop edges retrain}
    \label{fig:1by1bar_drop_cite}
  \end{subfigure}
  \hfill
  \begin{subfigure}[b]{0.3\textwidth}
    \includegraphics[width=\textwidth]{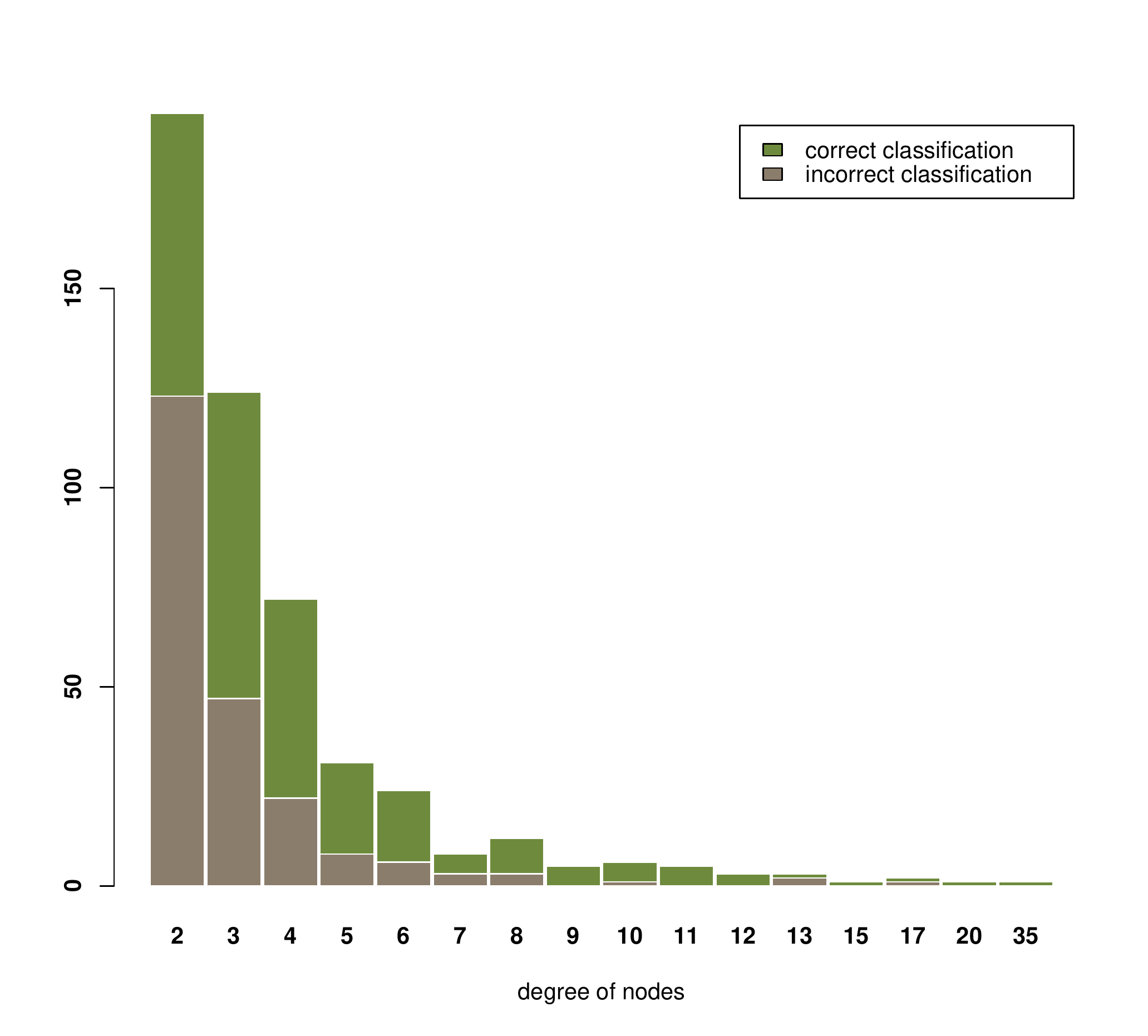}
    \caption{attack single nodes, Citeseer dataset, using adversarial training}
  \end{subfigure}
  \hfill
    \begin{subfigure}[b]{0.3\textwidth}
    \includegraphics[width=\textwidth]{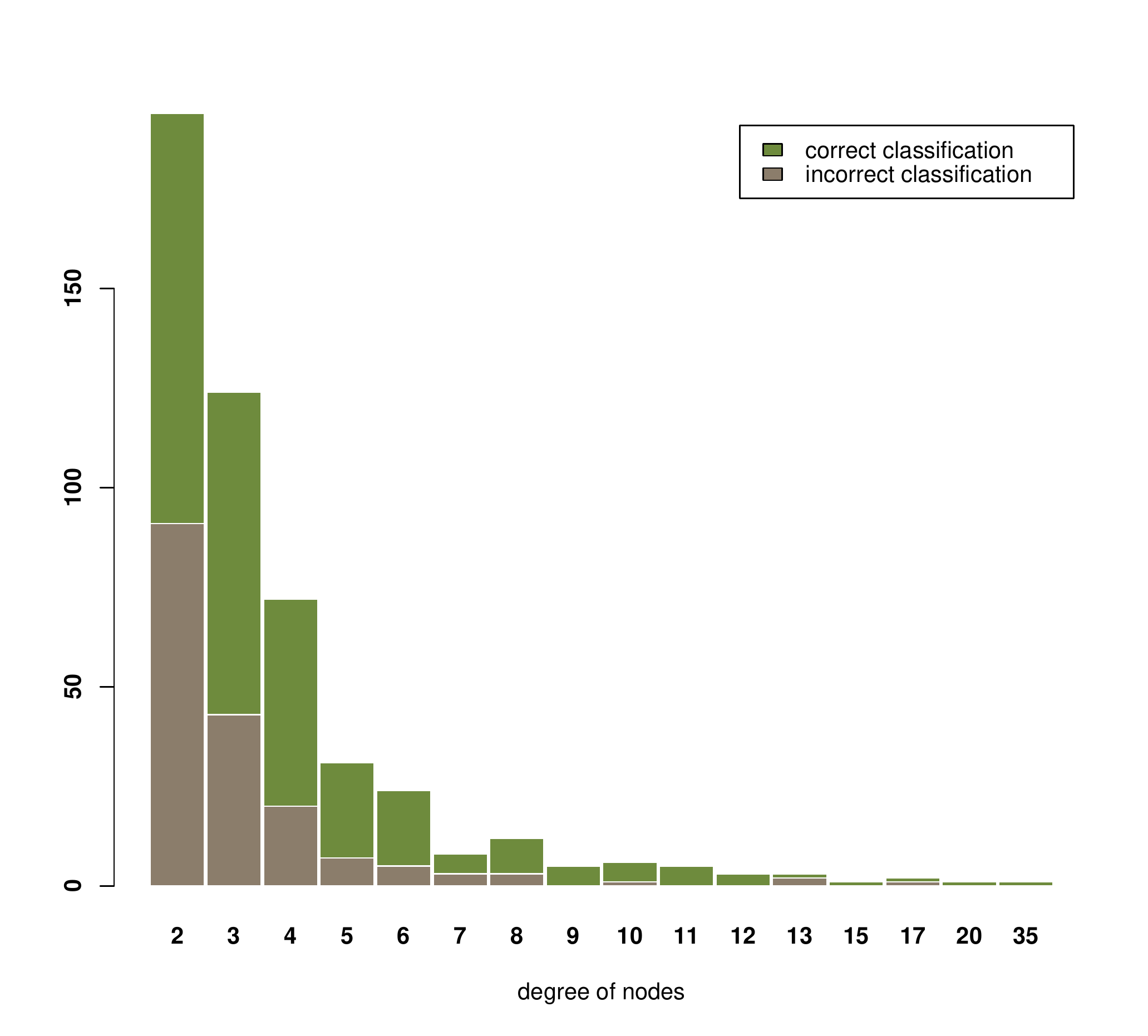}
    \caption{attack single nodes, Citeseer dataset, using GraphDefense}
  \end{subfigure}
  \hfill
  \begin{subfigure}[b]{0.3\textwidth}
   \includegraphics[width=\textwidth]{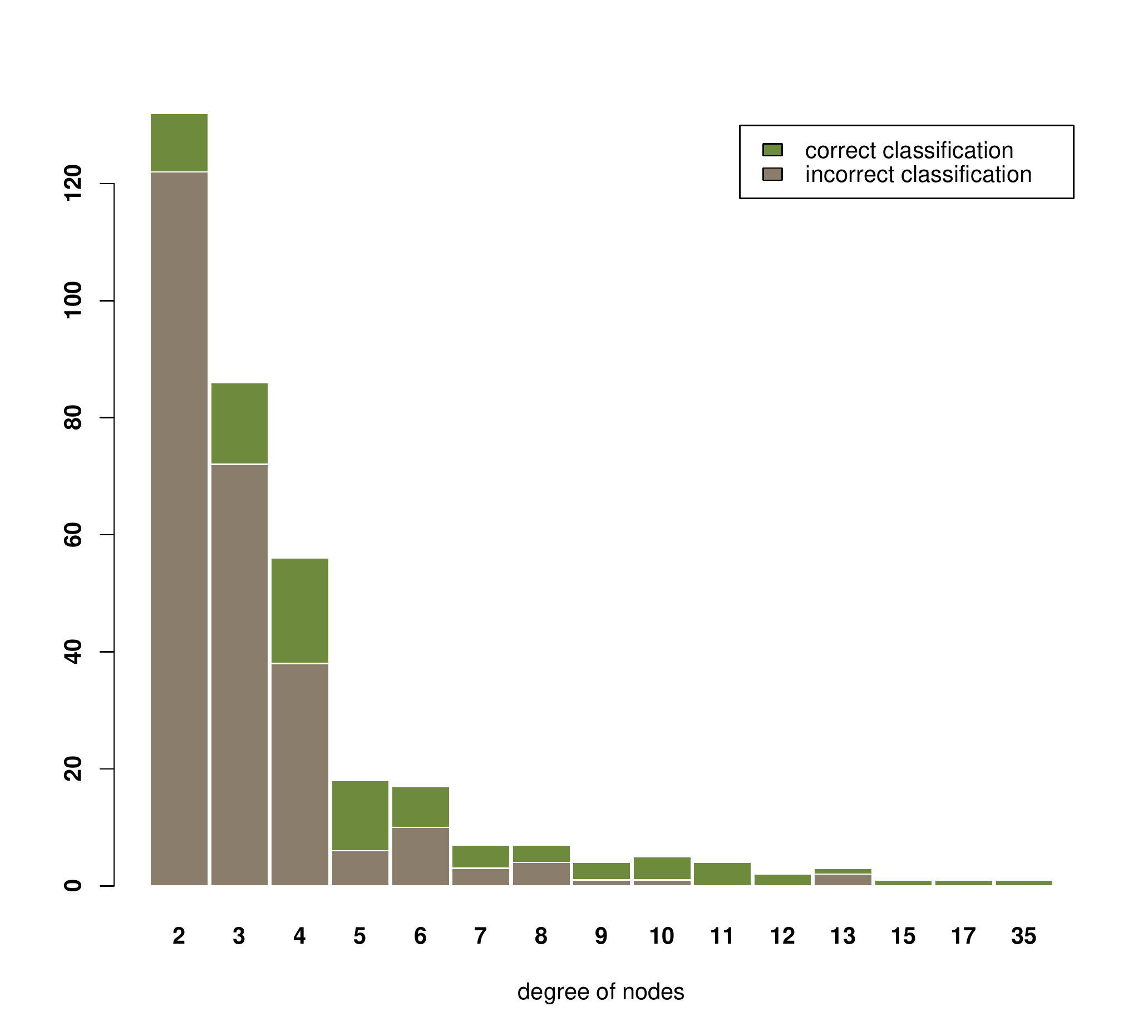}
   \caption{attack a group of 100 nodes, Citeseer dataset, clean model}
  \end{subfigure}
   \hfill
    \begin{subfigure}[b]{0.3\textwidth}
    \includegraphics[width=\textwidth]{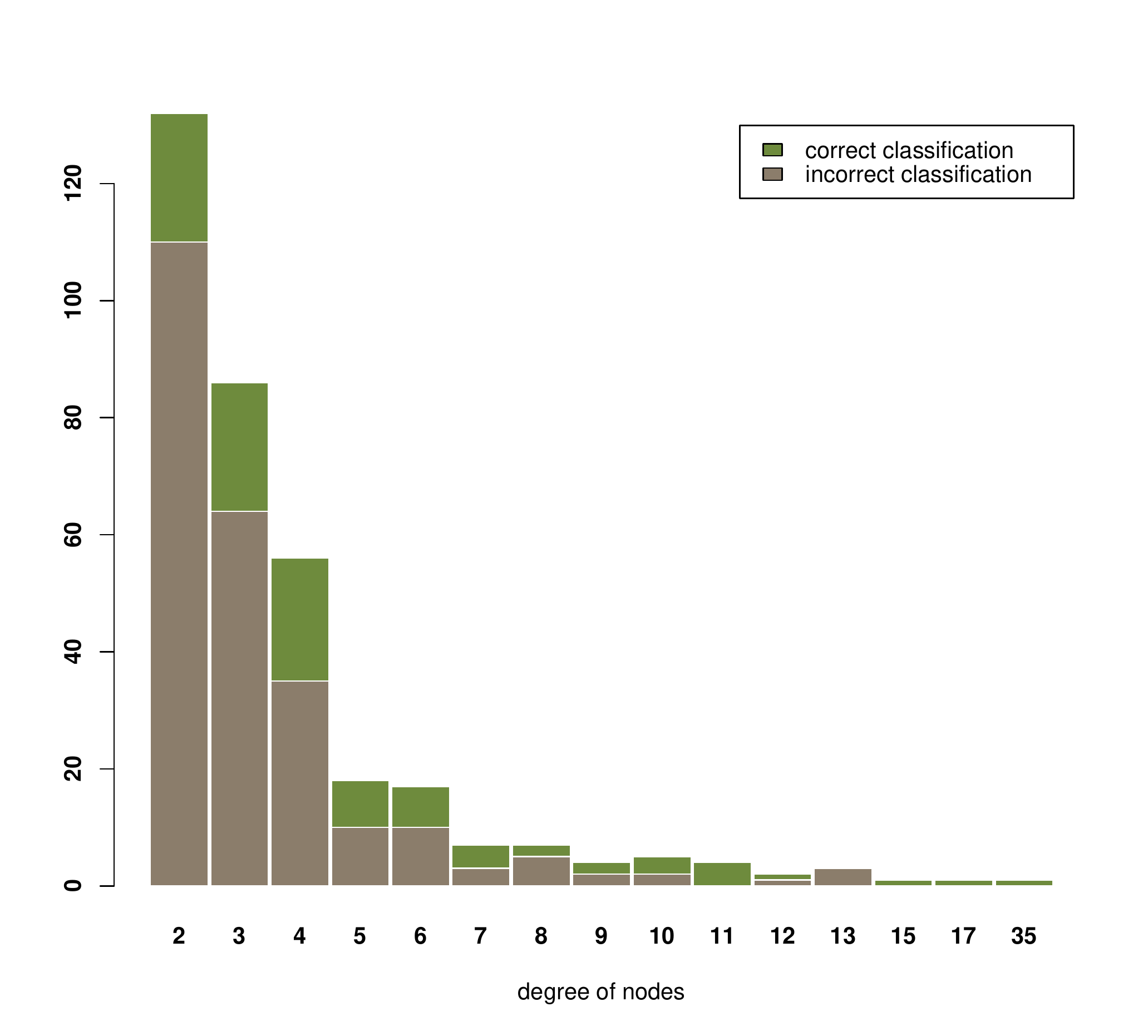}
    \caption{attack a group of 100 nodes, Citeseer dataset, using drop edges}
  \end{subfigure}
  
  \hfill
    \begin{subfigure}[b]{0.3\textwidth}
    \includegraphics[width=\textwidth]{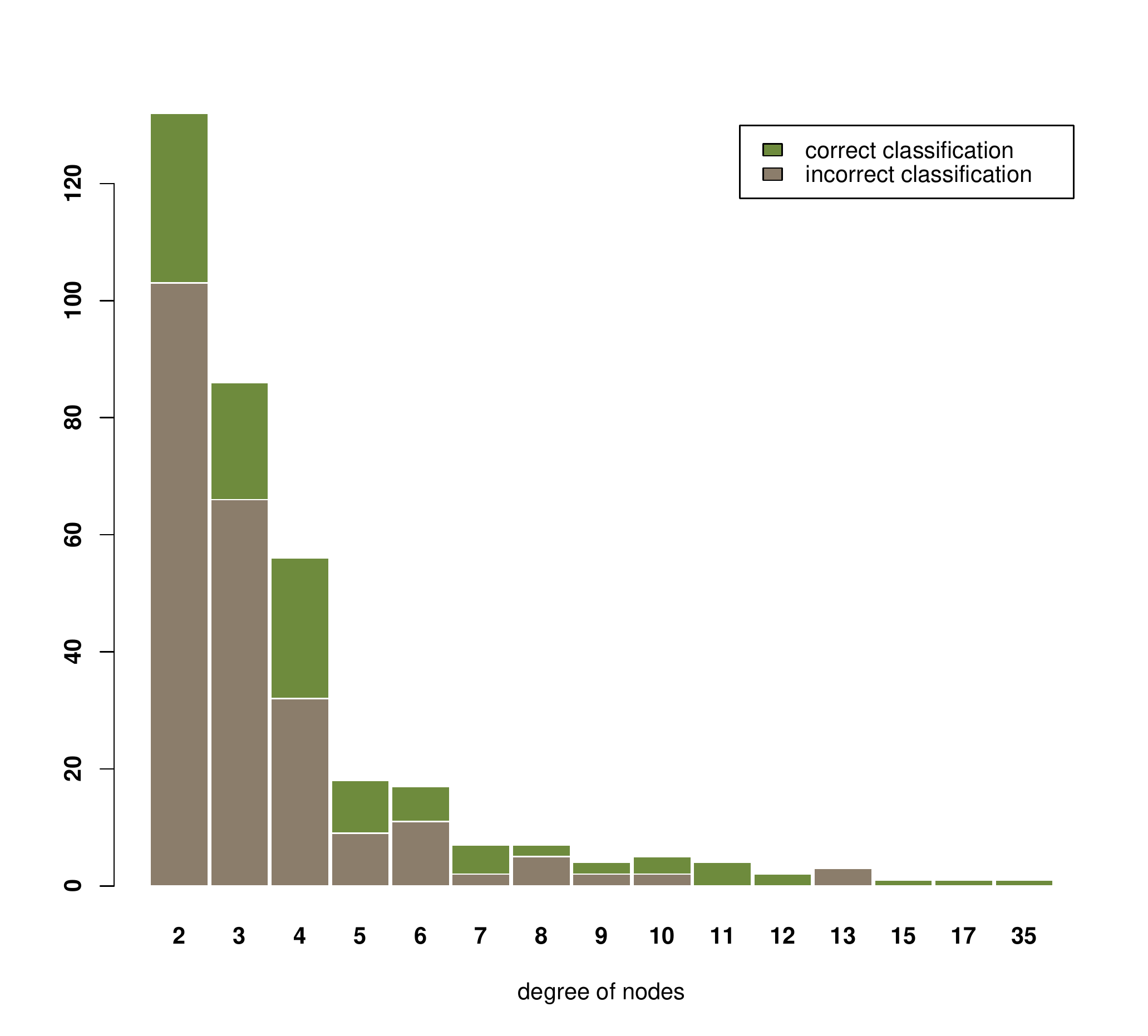}
    \caption{attack a group of 100 nodes, Citeseer dataset, using discrete adversarial training }
  \end{subfigure}
  \hfill
  \begin{subfigure}[b]{0.3\textwidth}
    \includegraphics[width=\textwidth]{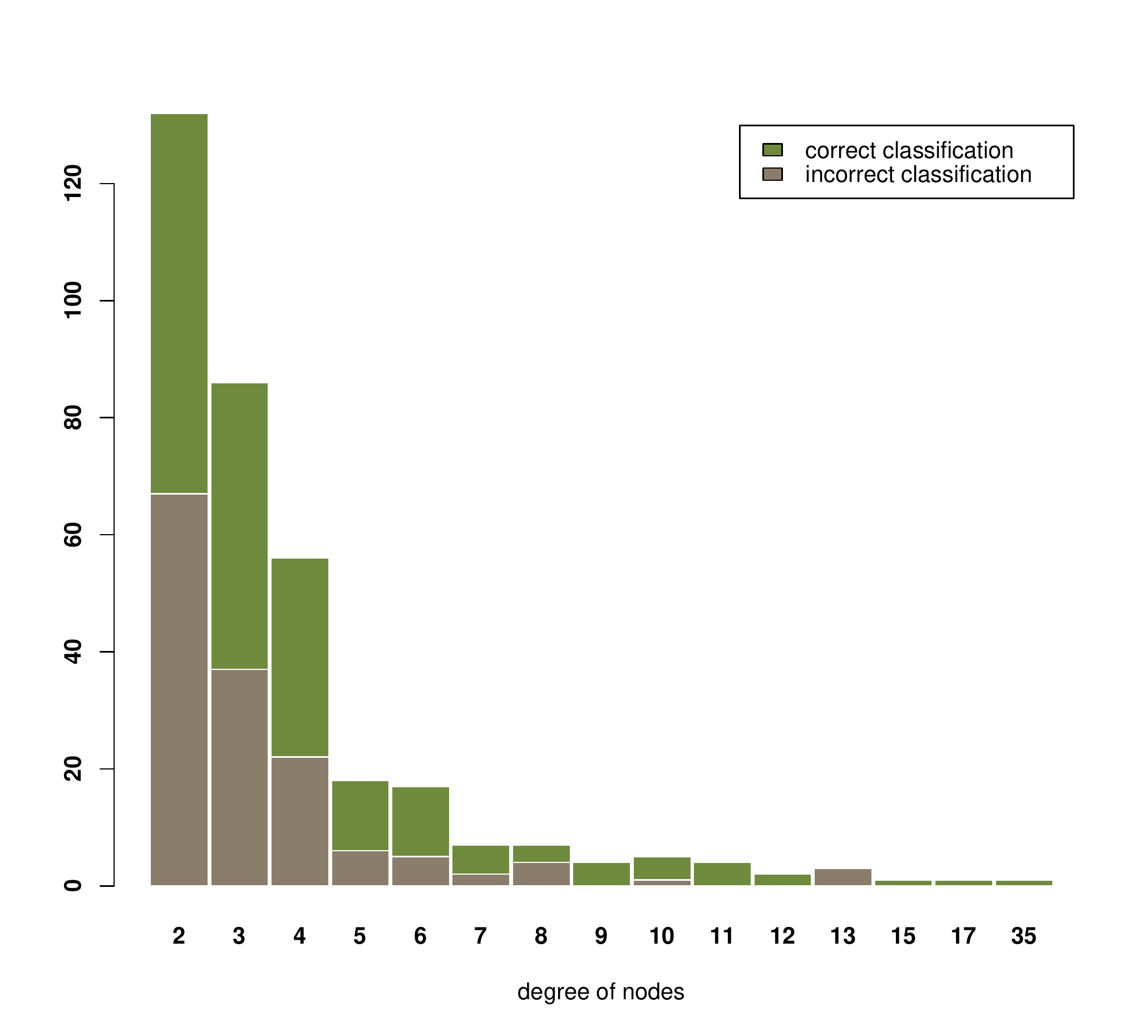}
    \caption {attack a group of 100 nodes, Citeseer dataset, using GraphDefense to defense}
  \end{subfigure}
\caption{ Attack Citseer dataset with different method. Green: correctly predicted nodes; Brown incorrectly predicted nodes.}
\label{fig:more_bars}
\end{figure*}

More Bar plots are listed in Figure \ref{fig:more_bars}, which shows each case how the accuracy changes before and after attacks. 

\subsection {Large scale and feature adversarial training}
\begin{figure}
    \centering
    \includegraphics[width=0.46\textwidth]{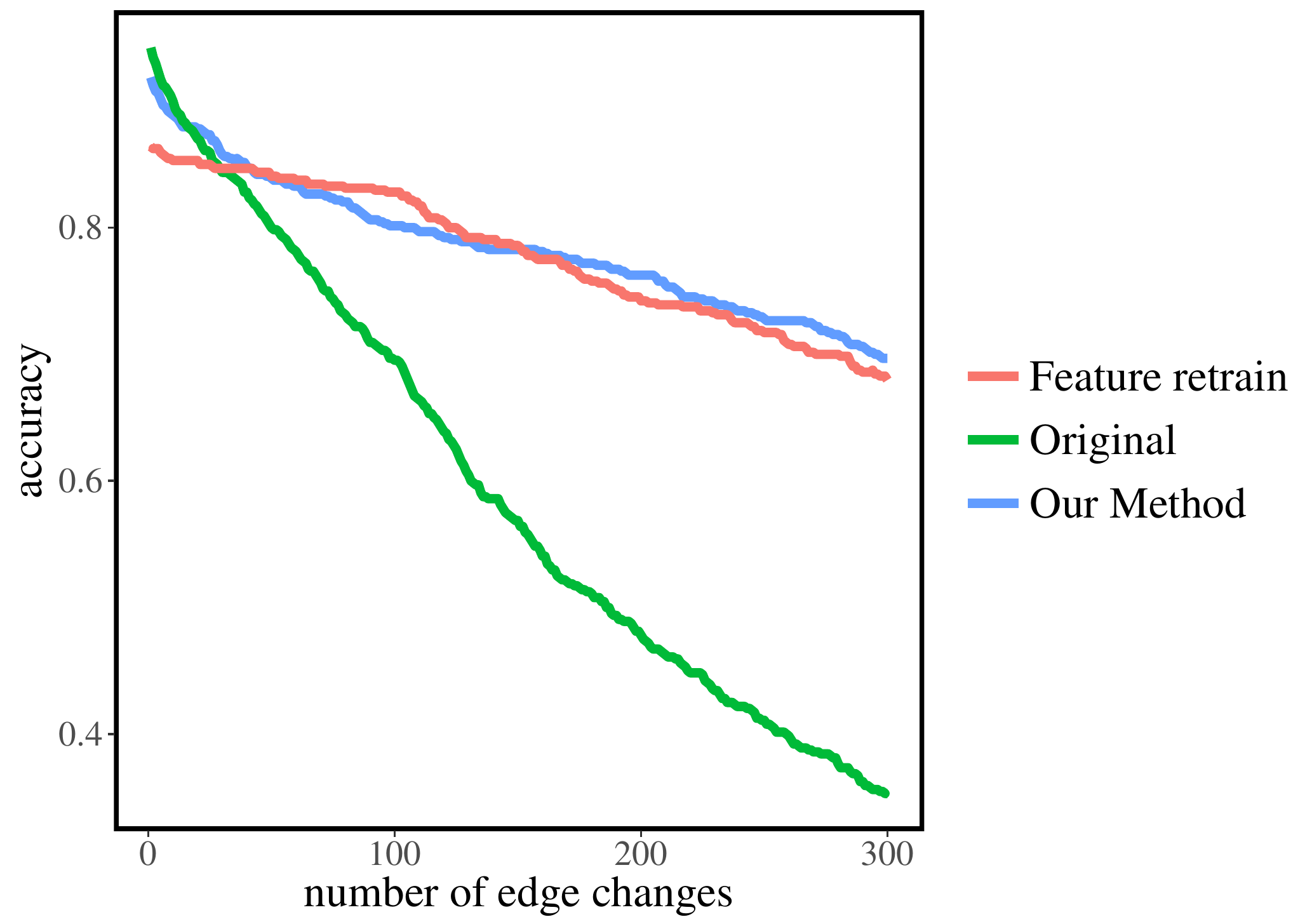}
    \caption{GraphSAGE accuracy after attacks for  adversarial training on features (Feature retrain) and on edges (our GraphDefense)  methods}
    \label{fig:graphsage}
\end{figure}

For large scale data, we use GraphSAGE compare with our GraphDefense method, discrete edges adversarial training, and adversarial training on features. We did not do a comparison with discrete adversarial training and dropping edges because previous experiments show they are far away behind our method. GraphSAGE is more difficult to attack, because there is a neighbourhood sampling function in these algorithms, directly adding or deleting edges on the original graphs method becomes less effective than on GCN. The reason is when training GraphSAGE (or other large scale graph neural networks), sampling neighbourhood could be view as dropping edges during training, \cite{pmlr-v80-dai18b} shows that use dropping edges while training  is a cheap method to increase the robustness of GCN. As a result, attacking GraphSAGE (or other large scale graph neural networks) is more difficult than attacking GCN. Also since attacking a single node by modifying only one edge is not a significant attack, in this part, we show attacking groups of 128 nodes instead. 

Because the Reddit dataset is an inductive dataset, using our framework Algorithm \ref{alg:framework} is important, otherwise, the adversarial training on the training dataset is very hard to transmit to the testing part through the edges, as the result, testing data will  remain vulnerable. 
\begin{table}
    \centering
    \caption{Result of average accuracy with different defense methods on GraphSAGE}
    \resizebox{0.46\textwidth}{!}{\begin{tabular} {|c|c|c|c|c|c|c|} 
        \hline
        &\textbf{before attack} & \textbf{50 edges} & \textbf{100 edges} & \textbf{150 edges} & \textbf{ 200 edges} & \textbf{300 edges}\\
        \hline
        Clean &0.9422  & 0.8 & 0.6953 & 0.5688 &  0.4781 &0.3531\\
        \hline
        Feature retrain X & 0.8641 &0.8406 & 0.8281& 0.7859 & 0.7422 &0.6797\\
        \hline
        Our method on A &0.9188 & 0.8391  & 0.8016 &  0.7828 & 0.7625 &0.6969\\
        \hline
    \end{tabular}}
    
    \label{tab:sage_defense}
\end{table}
Figure \ref{fig:graphsage} and Table \ref{tab:sage_defense} show attacking after different adversarial training methods. The result matches our claim in Section \ref{adv_in_feat}. Adversarial training in features has a similar result as in edges when facing attacks on edges, also adversarial training in features is faster than adversarial training in edges. The performance of adversarial training in features might be related to the data type. For example, Reddit dataset features are continuous while Cora and Citeseer are discrete. Although the result for adversarial training in feature for Cora is not as good as our GraphDefense method, it is still quite better than others, Cora dataset could remain 51 \% accuracy. 

\subsection{Parameter Sensitivity}
In this section, we will discuss the weight between adversarial examples and clean data during the adversarial training process in Algorithm \ref{alg:framework}.  
\begin{table}
\centering
\caption{Different number of Adversarial examples and clean examples during retrain process in Algorithm \ref{alg:framework}, Cora dataset.}
\resizebox{0.46\textwidth}{!} {\begin{tabular}{|c|c|c|c|c|}\hline
\backslashbox{\textbf{Adversarial}}{\textbf{Clean}}
&\makebox[3em]{100}&\makebox[3em]{200}&\makebox[3em]{300}&\makebox[3em]{400}\\\hline
100 & 0.634&0.692 & 0.616 & 0.622\\\hline
200 & 0.522 &0.54 & 0.555 & 0.55\\\hline
\end{tabular}}

\label{tab:para}

\end{table}

Table \ref{tab:para} shows that choosing an appropriate ratio between adversarial examples and clean examples during the adversarial training process is important. Too large portions of adversarial examples will cause lower accuracy, thus lead to bad performance after attacks.

\section{Conclusion}
In this paper, we propose a new defense algorithm call GraphDefense to improve the robustness of Graph Convolutional Networks against adversarial attacks on graph structures.
We further show that adversarial training on features is equivalent to adversarial training on graph structures, which could be used as a fast method of adversarial training without losing too much performance.
Our experimental results that our defense method successfully defense white-box graph structure attacks for not only small datasets but also large scale datasets with GraphSAGE \cite{graphsage} training. We also discuss what characteristics of defense methods are crucial to improve the robustness.

\bibliographystyle{plain}
\bibliography{references}

\end{document}